\documentclass[manuscript,screen]{acmart}
\usepackage{subfigure}
\usepackage{multirow}
\usepackage{enumitem}

\newcommand{\ruichao}[1]{\textcolor{black}{#1}}

\AtBeginDocument{%
  }

\setcopyright{acmcopyright}
\copyrightyear{2025}
\acmYear{2025}
\acmDOI{XXXXXXX.XXXXXXX}

\acmJournal{TIST}
\acmVolume{0}
\acmNumber{0}
\acmArticle{0}
\acmMonth{0}

\begin{document}

\title{
LLM-Enhanced Multiple Instance Learning for Joint Rumor and Stance Detection with Social Context Information}


\author{Ruichao Yang}
\email{csrcyang@comp.hkbu.edu.hk}
\affiliation{%
  \institution{Department of Computer Science, Hong Kong Baptist University, Hong Kong}
  \country{China}
}

\author{Jing Ma$^*$}
\affiliation{%
  \institution{Department of Computer Science, Hong Kong Baptist University, Hong Kong}
  \country{China}
}
\email{majing@comp.hkbu.edu.hk}
\thanks{$^*$Jing Ma is the corresponding author.}

\author{Wei Gao}
\affiliation{%
  \institution{School of Computing and Information Systems, Singapore Management University}
  \country{Singapore}
}
\email{weigao@smu.edu.sg}

\author{Hongzhan Lin}
\affiliation{%
  \institution{Department of Computer Science, Hong Kong Baptist University, Hong Kong}
  \country{China}
}
\email{cshzlin@comp.hkbu.edu.hk}

\renewcommand{\shortauthors}{R.C.Yang and J.Ma, et al.}

\begin{abstract}
The proliferation of misinformation, such as rumors on social media, has drawn significant attention, prompting various expressions of stance among users. Although rumor detection and stance detection are distinct tasks, they can complement each other. Rumors can be identified by cross-referencing stances in related posts, and stances are influenced by the nature of the rumor. However, existing stance detection methods often require post-level stance annotations, which are costly to obtain. We propose a novel LLM-enhanced MIL approach to jointly predict post stance and claim class labels, supervised solely by claim labels, using an undirected microblog propagation model. Our weakly supervised approach relies only on bag-level labels of claim veracity, aligning with multi-instance learning (MIL) principles. To achieve this, we transform the multi-class problem into multiple MIL-based binary classification problems. We then employ a discriminative attention layer to aggregate the outputs from these classifiers into finer-grained classes. 
Experiments conducted on three rumor datasets and two stance datasets demonstrate the effectiveness of our approach, highlighting strong connections between rumor veracity and expressed stances in responding posts. Our method shows promising performance in joint rumor and stance detection compared to the state-of-the-art methods.
\end{abstract}

\keywords{Multiple Instance Learning, Rumor Detection, Stance Detection, Propagation Structure, Hierarchical Attention Mechanism}

\received{7 June 2024}

\maketitle

\section{Introduction}

The rapid expansion of social networks has led to a proliferation of rumors, posing significant threats to the online community and causing detrimental consequences for individuals and society at large~\cite{10.1145/2870630}. For instance, during the COVID-19 pandemic, a false rumor claimed that ``magnetism will be generated in the body after the injection of coronavirus vaccine''\footnote{\url{https://www.bbc.com/news/av/57207134}}. This misinformation went viral and was shared millions of times on social media platforms, contributing to public vaccination hesitancy and delaying the establishment of herd immunity. 

Rumor detection typically aims to classify a piece of information, such as a given claim, into either a rumor or non-rumor~\cite{sharma2019combating,ding2022metadetector}. 
A noteworthy observation is that initially unverified posts, i.e., rumorous claims, may eventually be proven true, false, or a mixture of both. Therefore, a subtly different task from rumor detection is rumor verification, which involves estimating the specific veracity of a given claim related to a particular subject matter~\cite{gorrell2019semeval}. Rumor verification is, therefore, typically modeled as a multi-class rather than binary classification task. In this article, we do not make a strict distinction between these two and refer to both uniformly as rumor detection. 
Despite this difference, a crucial phenomenon is that posts related to rumors often trigger a wide range of mostly controversial stance expressions among online users. These stance signals have been found valuable for rumor detection in numerous previous approaches~\cite{ma2018detect,wei2019modeling}.
%
While some unsupervised stance detection methods have been suggested~\cite{kobbe2020unsupervised,allaway2020zero,li2023tts}, they tend to exhibit poor generalizability due to their reliance on manually crafted features or specific models pre-trained on certain topics or tasks, which are hardly transferable to free-form social media context. Consequently, there is a pressing need to develop a method that addresses the challenge of effective rumor-stance related tasks without assuming the availability of post-level stance labels. \looseness=-1

\begin{figure*}[t!]
\centering
\setlength{\abovecaptionskip}{0pt} 
\setlength{\belowcaptionskip}{0pt}
\includegraphics[width=3.5in]{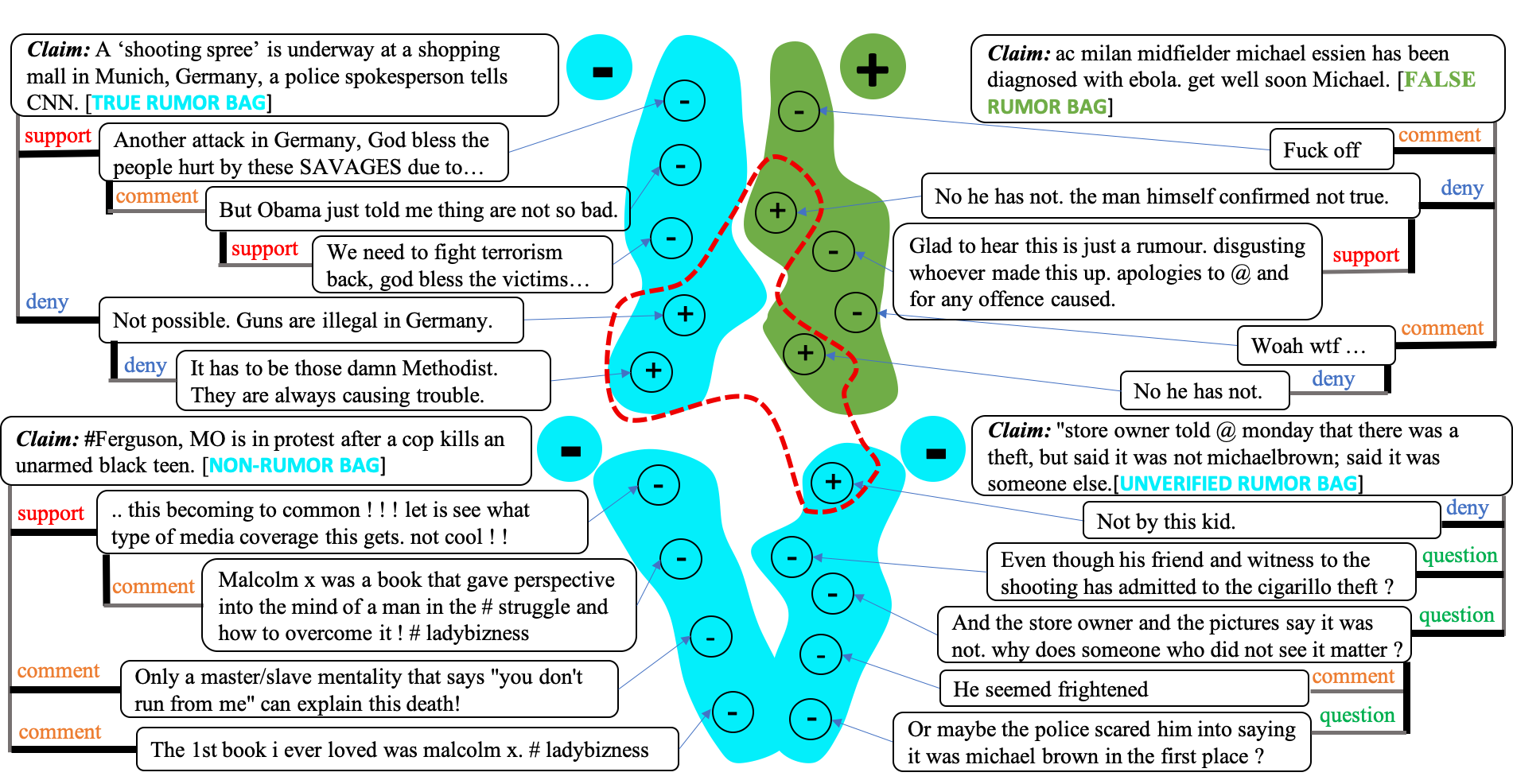}
\caption{An illustration of tree-based MIL binary classification for simultaneous rumor and stance detection.}
\vspace{-0.4cm}
\label{fig:intro}
\end{figure*}

To the best of our knowledge, no existing rumor-stance model has addressed the challenge of missing post stance labels while only having rumor veracity labels at the claim level, except for our recent work~\cite{DBLP:conf/sigir/YangMLG22}, which attempts to detect rumors and stances within a joint Multiple Instance Learning (MIL) framework using only claim veracity labels for supervision based on bottom-up/top-down propagation trees. Our approach is inspired by the observable patterns that correlate different stances with specific veracity classes of rumors. For instance, as shown in Figure~\ref{fig:intro}, a false rumor claim such as ``ac milan midfielder michael essien has been diagnosed with ebola'' sparks both commenting, denial, and questioning posts, while a true rumor claim such as ``A `shooting spree' is underway at a shopping mall'' mostly triggers supporting and comment posts among others. Meanwhile, the propagation structure of posts provides valuable insights into how a claim is disseminated and shaped among users over time~\cite{ma2020debunking,wei2019modeling}. Hence, the stances expressed in these structured responses contain useful signals for assessing the nuanced veracity of a rumorous event, such as true rumor, false rumor, unverified rumor, and non-rumor. Examining the veracity of rumors reveals that false rumors tend to attract more denial posts and exhibit a more intricate chain of interactions, such as ``$support \to deny$'' and ``$deny \to support$'', in contrast to true claims. Intuitively, we can deduce the stances of individual posts by initially determining the stances of key nodes directly, and then expanding to connected nodes using the distinctive edges in the response structure. \looseness=-1

This paper is a substantially extended version of our previous weakly supervised approach~\cite{DBLP:conf/sigir/YangMLG22}, based on MIL framework~\cite{foulds2010review}, which allows us to jointly detect rumors and stances using only the veracity labels of rumorous claims, referred to as bag-level (or event-level) annotation. Traditionally, an MIL-based classifier is used to classify individual instances (e.g., sentences) within a bag (e.g., document) and then deduce the bag-level prediction by aggregating the instance-level predictions, relying solely on bag-level annotations for weak supervision during training. 
However, the original MIL framework assumes binary and homogeneous class labels at both levels. In contrast, both rumor and stance detection typically deal with multiple class labels, with distinct differences between rumor and stance classes. The main challenge lies in establishing correlations between these two different sets of class labels and making both tasks learnable within the existing MIL framework. Another challenge is how to infer from one set of classes to another based on the feature representations learned from propagation trees, which embody the essential structure of social media posts.\looseness=-1 

To address these challenges, we first convert the multi-class classification problem into multiple MIL-based binary problems. One might wonder why multiple binary MIL-based problems could perform better in the joint detection of rumor and stance. Figure~\ref{fig:intro} illustrates the intuition of our joint prediction model. In this model, we treat the combination of False rumor and Deny stance (F-D pair) as the positive class (i.e., the target class) at both bag and instance levels, while considering all other classes as negative. Assuming that rumor veracity labels include four categories, and post stance labels also consist of four categories,
there exist various veracity-stance class pairs. Each veracity-stance pair can represent a target class for binary classification of claim veracity and post stance jointly. To synthesize the results from this set of veracity-stance binary classification models, we introduce a novel hierarchical attention mechanism. This serves two key purposes: (1) it aggregates the stances obtained from each binary MIL model to infer the veracity of a rumor claim, and (2) it combines the multiple binary results into a unified result that encompasses multiple classes in terms of veracity and stance. 


Furthermore, to strengthen post representation within the conversational threads, inspired by the success of the transformer network, we enhance the post encoder in our previous work~\cite{DBLP:conf/sigir/YangMLG22} by incorporating bottom-up/top-down tree transformers~\cite{ma2020debunking} to boost the post stance representation. Apart from this tree-based model, we further propose to leverage Large Language Models (LLMs) in our MIL-based framework to fully utilize the complex interactions between post pairs, overcoming the limitation of propagation directions. 
This is because a salient defect of our previous MIL-based approach~\cite{DBLP:conf/sigir/YangMLG22} is its oversimplified correlations for inferring veracity based on stances. To address the issue, we assume that LLMs can generate explanations for each post stance regarding the rumor type of the corresponding claim. 
While LLMs have shown impressive capabilities in various downstream tasks, their performance in rumor-stance tasks, particularly when considering social context information, remains understudied. Directly using LLMs to predict the stance of posts may introduce unpredictable noise, making it challenging to accurately predict rumor veracity. In light of this, our work focuses on using LLMs to generate reliable explanations for post stance. \ruichao{Another issue of our former MIL-based approach~\cite{DBLP:conf/sigir/YangMLG22} is that the model with a single propagation direction limits its broader contextual reasoning ability. To this end, we use LLM-generated explanations to enhance the representation of claims and posts, as well as the interaction between any two elements during the information dissemination process. In this way, our approach overcomes the limitation of the propagation direction, thereby improving performance in rumor-stance tasks within a unified framework.}

Consequently, this paper serves as a natural follow-up of our previous work~\cite{DBLP:conf/sigir/YangMLG22} with several substantial extensions: 1) We replace the original GRU-based bottom-up/top-down post encoder with a tree transformer-based framework to significantly enhance claim/post representation and robustness of the post encoder. 
2) We introduce an LLM-enhanced MIL method to overcome the limitations of existing bottom-up/top-down models, which are unable to capture the complex interactions between post pairs limited by propagation directions. 
\ruichao{3) We design a novel hierarchical stance tree attention mechanism enhanced by the LLM-generated stance explanation to make the binary class prediction on posts and claims, focusing on utilizing the most reliable posts and improving the explainability of the classification. 4) We develop a claim-explanation-guided attention aggregation mechanism that combines stance and rumor predictions from multiple binary MIL classifiers into a unified multi-class output. This mechanism offers a more robust way to combine multiple predictions by concentrating on the informative veracity-stance pairs, improving both rumor detection and stance results.
5) Conducting systematic experiments and analyses using datasets of different languages based on Twitter and Weibo. The results demonstrate that our proposed method achieves superior performance compared to state-of-the-art methods, indicating its generalize ability across different languages and social media platforms.}
Our paper makes the following contributions: \looseness=-1
\begin{itemize}
 \item
 \ruichao{We substantially extend the initial MIL-based weakly supervised method published in our recent work~\cite{DBLP:conf/sigir/YangMLG22}, 
by proposing a novel LLM-enhanced MIL model to enhance post encoding with stance explanations generated by LLM for improved representation of claims and posts, as well as the interaction between any two elements during the information dissemination process. Our approach overcomes the limitation of the propagation direction, thereby improving performance in rumor-stance tasks within a unified framework.}
 \item 
\ruichao{We design a  hierarchical stance tree attention mechanism to leverage LLM-generated stance explanations to improve rumor detection in binary classifiers. This novel mechanism integrates both post content and stance explanation, enabling more effective representations of complex inter-post relationships.}
\item 
\ruichao{We propose a new technique integrating claim explanations into the final attention process for multiple binary classifiers, enhancing the final prediction by incorporating the contextual insights from these explanations.}
 \item
 We achieve promising results on three widely used datasets based on Twitter and Weibo posts, demonstrating the effectiveness of our approach for both rumor and stance detection at the bag-level and post-level, respectively.
\end{itemize}

\section{Related Work}
\vspace{-0.1cm}
\vspace{-0.1cm}
\subsection{Rumor Detection} 

Early research on automatic rumor detection focused on pre-defined features or rules~\cite{middleton2016geoparsing} to train supervised classification models. To circumvent tedious feature engineering, subsequent studies proposed data-driven methods based on neural network models~\cite{lin2021rumor, sun2022ddgcn, lin2022detect} and multi-task learning frameworks~\cite{ma2018detect,wei2019modeling,luo2024joint}. Advanced approaches were then proposed to leverage propagation information~\cite{lu2020gcan, bian2020rumor,yu2020coupled,ma2021improving, lin2023zero}. 
In this study, we formulate our approach based on a propagation structured model, with a specific emphasis on deducing post-level stances using weak supervision with only claim-level annotation. Drawing inspiration from the efficacy of propagation information, our methods are designed based on graph structure for incorporating social context information.

\subsection{Stance Detection} 

Manual analysis of stance has revealed a close relationship between specific veracity categories and stances~\cite{mendoza2010twitter}. Subsequent studies explored various hand-crafted features~\cite{lukasik2016hawkes} 
for training stance detection models. More recently, deep neural networks have been employed for stance representation learning and classification~\cite{zhang2019stances,zhao2020pretrained,kochkina2017turing,wei2019modeling,liang2022jointcl}.
However, the aforementioned methods typically require a substantial stance corpus annotated at the post level for training. While several unsupervised models based on pre-defined models~\cite{kobbe2020unsupervised,allaway2020zero} and unsupervised structural embedding~\cite{DBLP:conf/aaai/RanJ23} for stance detection exist, their generalizability is consistently a critical concern due to their specific problem settings, reliance on manually crafted features.
In this paper, we propose a weakly supervised propagation model to concurrently classify rumor veracity and post stances on social media without the need for post stance labels. 

\subsection{Multiple Instance Learning (MIL)}

MIL is a form of weakly supervised learning designed to train a classifier with coarse-grained bag-level annotations to assign labels to fine-grained instances arranged in the bag~\cite{dietterich1997solving}. 
In recent years, MIL has proven effective and found successful applications in various Natural Language Processing (NLP) tasks~\cite{yang-etal-2023-wsdms}.
However, the original MIL framework is specifically designed for the binary classification of instances lacking complex structures, and bag-level labels are expected to conform to instance-level labels. 
In our work, we study an LLM-enhanced MIL framework to transform the multi-class problem into multiple binary classifiers, resolving the label incompatibility between bag and instance levels.

\subsection{Large Language Models (LLMs)}
Generative Pre-trained Transformers (GPTs) are a class of LLMs that have revolutionized artificial intelligence in recent years. 
The significant scientific advancements made by models such as ChatGPT~\cite{NEURIPS2020_1457c0d6}, LLaMA~\cite{touvron2023Llama}, and ALPACA~\cite{taori2023alpaca} have gained researchers' attention in the field of rumor detection and stance detection. ~\citet{yan2024enhancing} and ~\citet{yang-etal-2024-reinforcement} leveraged LLM to enhance the rumor detection, 
~\citet{lan2024stance} and ~\citet{gambini2024evaluating} utilized LLM for stance detection.
These methods primarily focus on directly predicting rumor or stance types using LLMs. 
In this paper, 
we leverage the generative capacity of LLMs to produce explanations for post stances towards claim veracity, aiming to enhance the reliability and explainability of these stances for final rumor detection. 

\section{Problem Statement} \label{sec:ps}
We define a rumor dataset as $\{\mathcal{C}\}$, where each instance $\mathcal{C}=(c,X,y)$ is a tuple consisting of a claim $c$, a conversation thread responding to it denoted as $X=(t_1, t_2, \cdots, t_T)$, and a veracity label $y$ of the claim. Note that although the posts are presented in chronological order, there are explicit connections such as response or repost relations among them, and message propagation paths (i.e., edges). 
We define the following two tasks: \looseness=-1

\begin{itemize}
\item \textbf{Stance Detection:} The task is to determine the post-level stance $p_i$ for a post $t_i$ expressed over the veracity of claim $c$. That is, $f: \{c, t_1, t_2, \dots t_T\} \to \{p_1, p_2, \cdots, p_T\}$, where $p_i$ is the stance label of $t_i$ that takes one of Support (S), Deny (D), Question (Q), or Comment (C)\footnote{Here the Comment is assigned to the posts that do not have clear orientations to the claim's veracity.}, and $\{c, t_1, t_2, \dots, t_T\}$ follows specific propagation structure (e.g., time sequence, tree,
or graph).
\item \textbf{Rumor Detection:} The task is to classify the claim $c$ on top of all the post stances as one of the four possible veracity labels $y\in \{\text{Non-rumor (N), True rumor (T), False rumor (F), Unverified rumor (U)}\}$. That is, $g: \{p_1, p_2, \cdots, p_T\} \to y$. In our approach, we hypothesize that the task can be modeled following the weighted collective assumption of MIL in the sense that the bag-level label relies more on the vital instances~\cite{foulds2010review}. 
\end{itemize}

\section{Weakly Supervised Propagation Model}\label{sec:Method}
The central concept of our approach is around detecting rumors through a cross-checking process by examining 
In this section, we will revisit our existing MIL-based methods~\cite{DBLP:conf/sigir/YangMLG22}, discuss their limitations, and then propose an LLM-enhanced method with MIL framework for joint rumor and stance detection. \looseness=-1


\subsection{Background} \label{sec:Background}
\begin{figure*}[t!]
\center
\setlength{\abovecaptionskip}{3pt} 
\setlength{\belowcaptionskip}{0pt}
\includegraphics[width=3.5in]{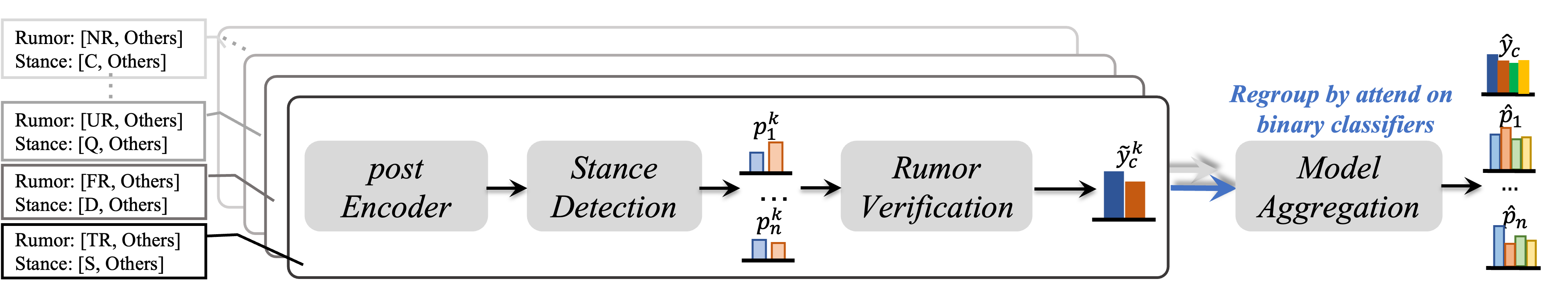}
\caption{An overview for our MIL-based weakly supervised framework.}
\vspace{-0.4cm}
\label{fig:TBmodel}
\end{figure*}

The original MIL framework is specifically designed for binary classification of instances lacking complex
structures, bag-level labels are expected to conform to instance-level labels~\cite {dietterich1997solving,foulds2010review}. Given the nuanced categories of stance and rumor veracity, we initially decompose the multi-class problem into multiple binary classification problems, and then we aggregate the predicted binary classes to form multi-way classes. Without the loss of generality, let us assume the number of rumor classes is $N_r$ and that of stance classes is $N_s$. As a result, there can be $K=N_r*N_s$ possible veracity-stance target class pairs, corresponding to $K$ binary classifiers weakly supervised for concurrently detecting rumor veracity and post stance. Figure~\ref{fig:TBmodel} provides an overview of our proposed method, which consists of the post encoder, stance detection, rumor detection, and model aggregation processes. 
However, existing MIL-based methods only use Gated Recurrent Unit (GRU)~\cite{cho2014properties} and bottom-up/top-down tree structure to represent posts, which cannot fully utilize interactions between posts at different positions. So, we intend to get a more nuanced understanding of each post via modeling the long-distance interactions among posts. \looseness=-1

An obvious defect of the above post representation learning process is that both bottom-up and top-down structures represent direct vertical responsive or reference relationships only, and it ignores the complex interactions between any pair of posts which could help post representation by gathering information from the other parallel informative posts. So, we disregard the direction of information flow and model the conversation topology among posts as an undirected tree, which retains the intrinsic characteristics (e.g., user stances) as described in the following section. As a result, the rumor-indicative features from the perspective of vertical and parallel perception can be fully captured. \looseness=-1
\vspace{-0.1cm}

\subsection{Post Encoder} \label{sec:GraphMIL}
The core idea of the post encoder in the weakly supervised propagation model lies in two main considerations. The first consideration is to utilize the complex interaction 
between a post and its parent (i.e., the influencing factors or prior context) as well as its child nodes (i.e., the downstream effects or spread). Through this way, we can leverage the structural dependencies to better understand 
how a post’s state is shaped by its position within the tree.
Another consideration is to resort to LLMs for 
generating explanations for each post’s stance. \looseness=-1

\begin{figure}[t!]
\centering
\setlength{\abovecaptionskip}{2pt} 
\setlength{\belowcaptionskip}{0pt}
\includegraphics[scale=0.28]{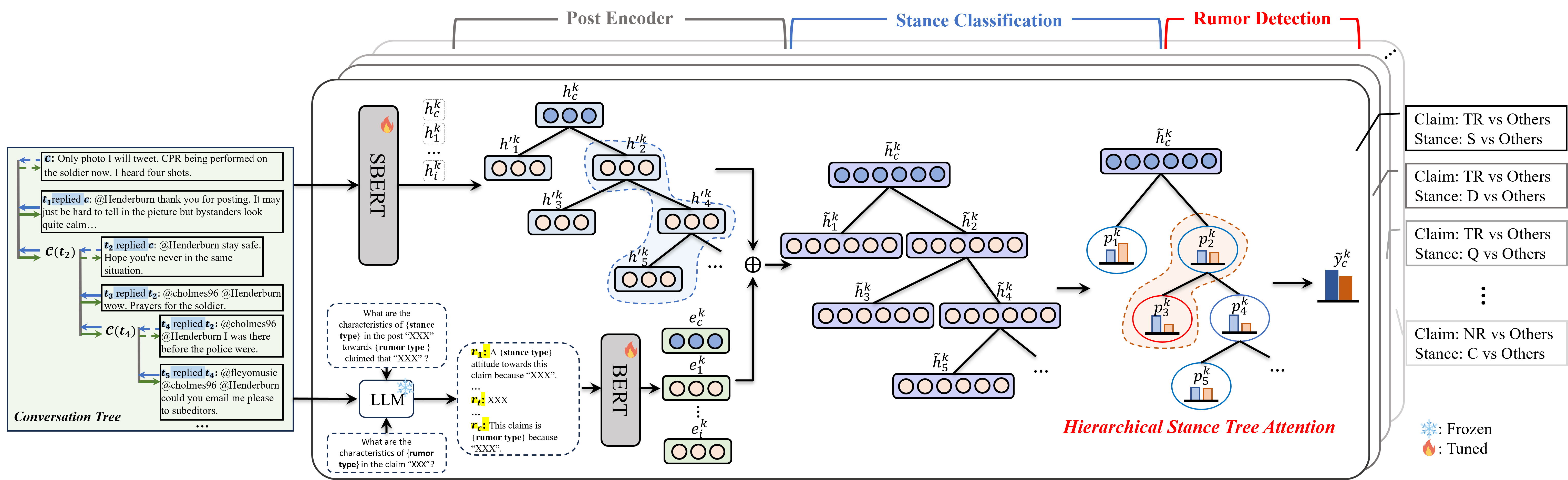}
\caption{A framework of LLM-enhanced weakly supervised propagation model. $\tilde{h}_i^k$ represents the final post encoding, $p_i^k$ denotes the stance probability of the post $t_i$, and $\tilde{y}_{c}^{k}$ represents the rumor $c$'s veracity probability given by classifier $k$.} 
\vspace{-0.4cm}
\label{fig:GraphModel}
\end{figure}

Our proposed post encoder 
as shown in Figure~\ref{fig:GraphModel}, where the input of the post encoder includes a claim, its corresponding posts, and the propagation structure. To provide the structure of the conversation tree to the LLM, we prepend a prefix to each post to explicitly present the structural information of the conversation thread before the encoding process, which is illustrated in the figure. For example, the prefixes are formatted as \textbf{"$t_1$ replied to $c$", "$t_2$ replied to $t_1$"}, and so on. These prefixes provide the model with a clear structural context, facilitating a better understanding of the relationships between posts.

In order to model the interactions between words at different positions to jointly attend to the tokens from different representation spaces, we transform $t_i$ to $h_i$ and $c$ to $h_c$ with the Sentence-BERT~\cite{reimers2019sentence}\footnote{Here we opted for Sentence-BERT as it yielded the most superior experimental results compared with other transformer architectures. Without loss of generality, Sentence-BERT can be replaced with other language models such as ELMO~\cite{peters2018deep}, Roberta~\cite{liu2019roberta}, and BERT~\cite{devlin2019bert}.}. Let a post of word sequence $t_i=\{w_{1} w_{2} \cdots w_{|t_i|}\}$, where $w_{j} \in \mathbb{R}^d$ is a $d$-dimensional vector that can be initialized with pre-trained word embeddings. We represent the claim and its related posts using Sentence-BERT~\cite{reimers2019sentence} that maps the sequence into a fixed-size vector: 
\begin{equation}\label{equ:Token-levelRepre}
\small
\begin{split}
    & h_{c} =SBERT(w_1, w_2, \cdots, w_{|c|}, \theta_{c}) \\
    & h_{i} =SBERT(w_1, w_2, \cdots, w_{|t_i|}, \theta_{X})
\end{split}
\end{equation}
where $SBERT(\cdot)$ denotes the standard Sentence-BERT transition equations, $|\cdot|$ is the number of words, $w_{|c|}$ and $w_{|t_i|}$ are the last word of $c$ and $t_i$, respectively. And $\theta_{c}$ and $\theta_{X}$ contain all the parameters in the post encoder.

\subsubsection{Post-level Representation}
We model the interaction among posts as an undirected tree to eliminate direction restrictions, allowing potential information flow between any pair of posts.
For example, the child nodes reflect the attitude tendency towards the parent node, and the parent node guides the attitude tendency of the child node. 
Let $\mathcal{M}(i)$ denote the node set connected to $t_i$ (except the source node $c$). The post-level representation can be updated as follows: \looseness=-1
\begin{equation}\label{equ:PostLevelRepre}
\small
\begin{split}
    & \alpha_{ij}^{k} = \frac{exp({h}_{j}^{k} \cdot {h_i^{k}}^\top)}{\sum_{j \in \mathcal{M}(i)}exp({h}_{j}^{k} \cdot {h_i^{k}}^\top)} \\
    & {h'}_{i}^k = \rho {h}_{i}^k + (1-\rho) \sum_{j \in \mathcal{M}(i)}\alpha_{ij}^{k} \cdot h_{j}^{k}
\end{split}
\end{equation}
where $\alpha_{ij}^{k}$ denotes the attention coefficient of each neighbor node $t_j$ of the node $t_i$, $h_{j}^{k}$ denotes the post embedding obtained in Equation~\ref{equ:Token-levelRepre}, and $\rho$ represents the retention ratio of the current node. We empirically set $\rho$ in our experiments.

\subsubsection{Explanation-level Representation}
The shallow text representation of posts above is insufficient to fully encode the underlying context and nuanced attitudes for accurately identifying rumor and stance types. Previous studies indicate that LLMs can capture intricate relationships among words, context, etc., to generate deeper semantic representations of text for rumor detection~\cite{yan2024enhancing}. 
Therefore, we utilize ChatGPT\footnote{ChatGPT can be replaced by other 
LLMs such as Llama 2~\cite{touvron2023Llama}, OPT~\cite{zhang2022opt}, BLOOM~\cite{le2023bloom}, and Claude~\cite{caruccio2024claude} for explanation generation.} to generate contextually-aware explanations, providing insights into the complex relationships between posts and their associated rumor or stance types. 
For example, we craft the following prompt to guide the LLM to generate stance explanations for a post: 

\textit{What are the characteristics of "support stance" in the post "$t_4$ replied $t_2$: @cholmes96 @Henderburn I was there before the police were.", towards "true rumor" claimed that "$c$: Only photo I will tweet. CPR being performed on the soldier now. I heard four shots." ?}

This example is designed to generate 
stance explanations for a post with a supportive stance towards a true rumor. The stance type, post content, rumor type, and claim content in the quotation marks can be directly replaced for the corresponding information of a different input. 
Based on the post content and stance explanations, we can further prompt the LLM to generate claim-type explanations for binary models aggregation. Here is an example prompt:

\textit{What are the characteristics of "Ture rumor" in the claim "$c$: Only photo I will tweet. CPR being performed on the soldier now. I heard four shots." ?}

Since the generated rationales are typically longer than a post, we encode each explanation $R_i$ with BERT~\cite{devlin2019bert} rather than Sentence-BERT, which yields:
\begin{equation}\label{equ:ExplanationEncode}
\small
    e_c^k, e_i^k = BERT(r_c, r_1, \cdots, r_{|{R}_{i}|}, \theta_{r}^k)
\end{equation}
where $|\cdot|$ and $r_{|{R}_{i}|}$ denote the number of words and the last word of explanation $R_i$, respectively, $\theta_{r}$ represents the parameters in BERT encoder. Then we concatenate the post-level and explanation-level representations to get the final post representation for binary  post prediction, which yields:
\begin{equation}\label{equ:FinalPostRepre}
\small
\begin{split}
    & \tilde{h}_{c}^{k} = {h}_{c}^{k} \oplus {e}_{c}^{k} \\
    & \tilde{h}_{i}^{k} = {h'}_{i}^{k} \oplus {e}_{i}^{k} \\
\end{split} 
\end{equation}
where $h_c^k$ is the claim embedding obtained in Equation~\ref{equ:Token-levelRepre}, ${h'}_{i}^{k}$ denotes the post-level representation obtained in Equation~\ref{equ:PostLevelRepre}, ${e}_{i}^{k}$ represents the explanation-level representation obtained in Equation~\ref{equ:ExplanationEncode}.

\subsection{MIL-based Binary Stance Classification}\label{sec:mil-binary}

In our LLM-enhanced weakly supervised model, each binary classifier follows a variant of standard MIL~\cite{angelidis2018multiple}, where the target class is treated as positive and the other classes as negative. For classifier $k$, to better capture the stance of each post towards the claim, we couple each claim-post pair $(c, t_i)$ and use a fully connected softmax layer to predict the binary stance probability distribution of $t_i$ towards the claim vector $h_c^k$ based on classifier $k$: 
\begin{equation}\label{equ:postdistributions}
\small
p_{i}^{k} = softmax(W_{1}^{k} \tilde{h}_{i}^{k} + W_{2}^{k} {h'}_{c}^{k} + b^k_o)
\end{equation}
where $W_{*}^{k}$ are trainable weights, $\tilde{h}_{i}^{k}$ and ${h'}_{c}^{k}$ are obtained in Equation~\ref{equ:FinalPostRepre} and Equation~\ref{equ:PostLevelRepre}, respectively. $b^k_o$ are the weights and bias. The probability values $\{p_{i}^{k}\} (j=1,2,\cdots,T)$ can be considered as the stance-indicative probabilities of all the nodes inferred from the claim veracity. 

The key intuition is that the binarization of the classes can be used to distinguish the salience of mappings between rumor and stance classes. So, the model can focus more on a salient subset of veracity-stance target class pairs such as F-D, T-S, U-Q, and N-C, embodying that close veracity-stance correspondences~\cite{mendoza2010twitter} will outweigh other possible pairs by using an attentive process to automatically adjust the weights of the binary models corresponding to the pairs. We will present an aggregation method in Section~\ref{sec:aggregation} to combine the predictions of binary models based on the learned weights for the final multi-class prediction.  \looseness=-1

\vspace{-0.1mm}
\subsection{MIL-based Binary Rumor Classification}

To infer claim veracity based on the posts' stances, we propose to attend over evidential nodes within the stance tree. For this purpose, we design a \textit{Hierarchical Stance Tree Attention Mechanism} to aggregate the stances, which consists of \textit{Local Attention} and \textit{Stance Explanation-guided Global Attention} Mechanism. \looseness=-1
\vspace{-0.1cm}

\subsubsection{Local Attention} The Local Attention is designed to simulate the local interaction and capture the fine-grained stance tendencies in the conversation tree. For example, the child nodes tend to reflect the attitude towards their parent node, and the parent node tends to guide the attitude of its child nodes. So, we aim to integrate each node's stance with its parent and child nodes to better capture the local context. 
Specifically, 
let $\mathcal{P}(i)$ denote the node set connected to $t_i$ (except the source node $c$). The stance probability distribution of $t_i$ can be updated as follows:
\begin{equation}\label{equ:GraphLocalAtt}
\small
    \begin{split}    
    & \alpha_{ij}^{k} = \frac{exp(\tilde{h}_{j}^{k} \cdot \tilde{h}_{i}^{{k}^\top)}}{\sum_{j \in \mathcal{M}(i)}exp(\tilde{h}_{j}^{k} \cdot \tilde{h}_{i}^{{k}^\top)}} \\
    & \tilde{p}_{i}^k = \lambda {p}_{i}^k + (1-\lambda) \sum_{j \in \mathcal{M}(i)}\alpha_{ij}^{k} \cdot {p}_{j}^{k}
    \end{split}
\end{equation}
where $\alpha_{ij}^{k}$ denotes the attention coefficient of each neighbor node $t_j$ of the node $t_i$, $\tilde{p}_{i}$ denotes the updated stance probability of node $t_i$,
and $\lambda$ represents the retention ratio of the current node. We empirically set $\lambda$ in our experiments. \looseness=-1

\subsubsection{Stance Explanation-guided Global Attention} 
Intuitively, LLM can generate explanations that align local stance interactions (e.g., post-to-claim or post-to-post) with bag-level claim veracity. So, we design global attention to differentiate the importance of different local stances enhanced by explanations for rumor veracity inference. 
Specifically, we first compute the attention coefficient that indicates the importance of each node towards the claim based on explanation-level representation. And then aggregate the stances according to the attention coefficients. 
This yields: \looseness=-1
\begin{equation}\label{equ:GraphRumorPredict}
\small
\begin{split}
    & \delta_{i}^{k} = \frac{exp({e}_{i}^{k} \cdot {h_c^{k}}^\top)}{\sum_{i=1}^{n} exp({e}_{i}^{k} \cdot {h_c^{k}}^\top)} \\
    & \tilde{y}_{c}^{k} = \sum_{i=1}^{n}\delta_{i}^{k} \cdot \tilde{p}_{i}^{k}
\end{split}
\end{equation}
where $\delta_{i}^{k}$ denotes the attention coefficient of each node, and $\tilde{y}_{c}^{k}$ is the predicted probability of rumor veracity. 
\looseness=-1

\subsection{Binary Models Aggregation} \label{sec:aggregation}

\begin{figure}[t!]
\centering
\vspace{-0.3cm}
\setlength{\abovecaptionskip}{0pt} 
\setlength{\belowcaptionskip}{0pt}
\includegraphics[width=3.6in]{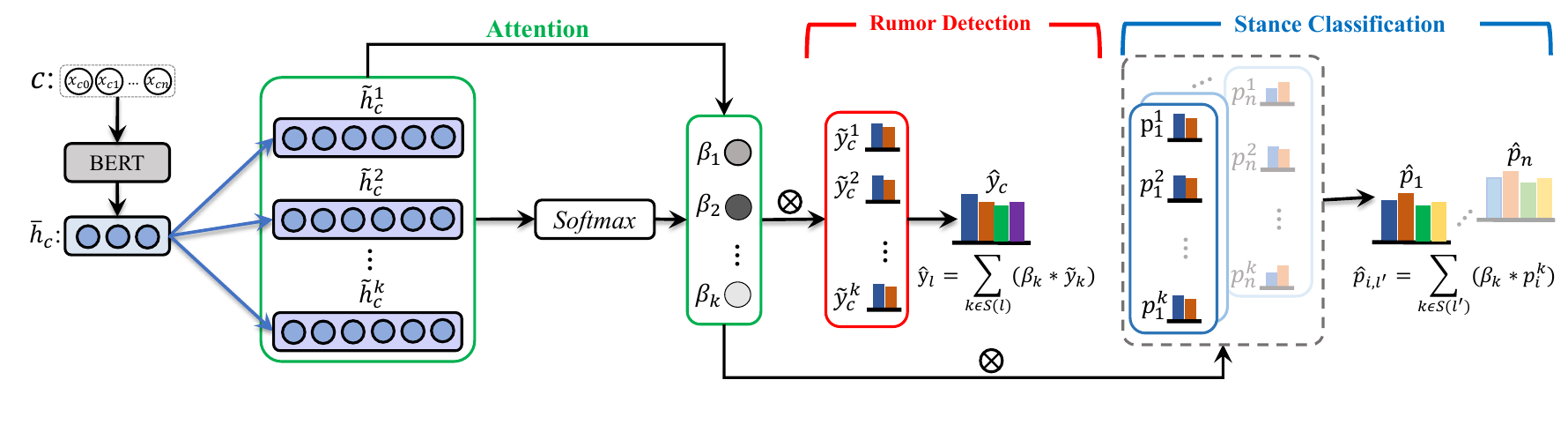}
\caption{The aggregation model for predicting fine-grained claim veracity and post stance, where $\beta_k$ is the attention score obtained from Equation~\ref{equ:betaattention}, $\tilde{y}_{k}$ and $p_i^k$ denote binary rumor and stance probabilities computed by the $k$-th classifier, respectively.}\label{fig:model_agg}
\vspace{-0.3cm}
\end{figure}

It is intuitive to contend that each binary classifier contributes differently to the final multi-class prediction, due to the different capacity of each model capturing the veracity-stance correlation. Therefore, we design a claim explanation-guided attention mechanism to attend over the binary classifiers 
as shown in Figure~\ref{fig:model_agg}. This yields:
\begin{equation}\label{equ:betaattention}
\small
    \begin{split}
    & \bar{h}_c = BERT(w_1, w_2, \cdots, w_{|c|}, \bar{\theta}_c)\\
    & \beta_{k} = \frac{exp({\bar{h}_c}\cdot \tilde{h}_{c}^{{k}^\top})}{\sum_{k}exp(\bar{h}_c \cdot \tilde{h}_{c}^{{k}^\top})}
    \end{split}
\end{equation}
where BERT represents the standard BERT transition equation, $\bar{h}_c$ denotes the global representation of the claim $c$, $|\cdot|$ is the number of words, $ w_{|c|}$is the last word of $c$, $\bar{\theta}_c$ denotes all the parameters in the claim encoder\footnote{Note that the BERT transition equation has a similar yet independent set of parameters with Equation~\ref{equ:ExplanationEncode}.}, and $\tilde{h}_c^k$ is the claim representation fused with claim explanation obtained from the $k$-th classifier (see Equation~\ref{equ:FinalPostRepre}). 

Finally, on top of the obtained attention scores of binary models, we use a weighted aggregation method to obtain the prediction of \emph{stance classification} and \emph{rumor veracity} by summing over $K=N_r*N_s$ attention coefficients and $K$ classifiers outputs:\looseness=-1

\textbf{Stance Detection.} 
For each post $t_i$ under claim $c$, we obtain $K$ probability scores from all the binary classifiers, i.e., $\{p_{i}^{1}, p_{i}^{2}, \cdots, p_{i}^{K}\}$. We utilize an attention-weighted sum to get the final post-level stance prediction $\hat{p}_{i}$. We regroup all the binary stance classifiers with the same target stance class $l_s \in \{S, D, Q, C\}$ into one set and then compute the final stance probability by:
\begin{equation}\label{equ:finalpostpredict}
\small
  \hat{p}_{i,l_s} = \sum_{k \in \mathcal{I}(l_s)}\beta_{k} \cdot p_{i}^{k} 
\end{equation}
where $\mathcal{I}(l_s)$ represents the indicator set of the binary classifiers with $l_s$ as the target stance class, $p_{i}^{k}$ is the predicted stance probability of the post $t_i$ given by classifier $k$, and therefore $\hat{p}_{i,l_s}$ indicates the probability that the post $t_i$ is classified as stance $l_s$. 
Note that these binary classifiers of the same target stance class are different models because each classifier is supervised by a different target veracity label of the claim. Thus, the final probability distribution over all the stances can be obtained as $\hat{p}_i=[\hat{p}_i^S, \hat{p}_i^D, \hat{p}_i^Q, \hat{p}_i^C]$. 

\textbf{Rumor Detection.} We regroup all the binary claim veracity classifiers and put the classifiers with the same rumor class label $l_c\in\{N,T,F,U\}$ into one set. And then the claim veracity probability can be computed similarly to the weighted sum of all the outputs from binary rumor veracity classifiers: 
\begin{equation}\label{equ:finalclaimpredict}
\small
  \hat{y}_{c,l_c} = \sum_{k \in \mathcal{I}(l_c)}\beta_{k} \cdot \tilde{y}_c^k 
\end{equation}
where $\mathcal{I}(l_c)$ is the indicator set of the binary classifiers with $l_c$ as the target veracity class and $\tilde{y}_c^k$ is the predicted binary veracity class probability of the claim $c$ given by the classifier $k$. Note that the binary veracity classifiers with the same target veracity class are distinguishable by the different target stance classes they are associated with, which give different stance probability predictions. Thus, the final probability distribution over the rumor veracity classes can be represented as $\hat{y}_c=[\hat{y}_c^N, \hat{y}_c^T, \hat{y}_c^F, \hat{y}_c^U]$.\looseness=-1


\section{Model Training}

To train each binary classifier, we transform the finer-grained rumor and stance labels into binary labels for ground-truth representation. For example, for the classifier with [T, S] (i.e., True-Support) veracity-stance pair as the target, the ground-truth veracity label of claim $y$ is represented as either `T' or `others', and the model outputs the stance for each post represented by a probability belonging to target class `S'. A variant of this similar setting is applicable to all the binary classifiers with different target classes. This yields the way to obtain our binary ground truth:
\begin{equation}\label{equ:WeakClassifersLabelBags}
\small
  y^{k}=
\begin{cases}
1& \text{if the target of classifier $k$ is the same as $y$}\\
0& \text{others}
\end{cases}  
\end{equation}
where $y \in \{N,T,F,U\}$ is defined for the rumor detection task, and the ground truth refers to the veracity label of claims instead of stance due to the unavailability of the label at the post level.

\textbf{Binary MIL-based Classifiers Training.} 
All the binary classifiers are trained to minimize the negative log-likelihood between the predicted distribution of a given claim and that of the ground truth:
\begin{equation}\label{equ:lossfunction}
\small
L_{bin} = -\sum_{k=1}^{K}\sum_{c=1}^{C} y_c^k*\log \hat{y}_c^k + (1-y_c^k)*\log(1-\hat{y}_c^k)
\end{equation}
where $y_c^k \in \mathbb [0, 1]$ indicating the ground truth of the $c$-th claim obtained in Equation~\ref{equ:WeakClassifersLabelBags}, $\hat{y}_c^k$ is the predicted probability for the $c$-th claim in classifier $k$, $C$ is the total number of claims, and $K$ is the number of binary classifiers. 

\textbf{Aggregation Model Training.} 
To train the aggregation model, we again utilize the negative log-likelihood loss function to minimize the distribution of claim prediction and that of the ground truth:\looseness=-1

\begin{equation}\label{equ:finallossfunction}
\small
L_{agg} = -\sum_{n=1}^{N} \sum_{m=1}^{M} y_{m,n}*\log \hat{y}_{m,n} + (1-y_{m,n})*\log(1-\hat{y}_{m,n})
\end{equation}
where $y_{m,n} \in [0, 1]$ is the binary value indicating if the ground-truth rumor veracity class of the claim $c_n$ is $m$, $\hat{y}_{m,n}$ is the predicted probability the claim $c_n$ belonging to class $m$ by Equation~\ref{equ:finalclaimpredict}, and $M$ is the number of rumor veracity classes.\looseness=-1

\textbf{Training Settings.}
During training, all the model parameters are updated by back-propagation with Adam~\cite{DBLP:journals/corr/KingmaB14} for speeding up convergence. We use pre-trained GloVe Wikipedia 6B word embeddings~\cite{pennington2014glove} to represent input words, set the model dimension $d$ to 300, feed-forward network dimension to $600$, and empirically initialize the learning rate as 0.001. The training process ends when the loss value converges or the maximum epoch number is met, for which We set the number of the maximum epochs as 200. 
We set the retention ratio of the current node $\lambda=0.7$. We train the aggregation model after all the binary classifiers are fully trained. When training the aggregation model, only the parameter $\theta_a$ is updated while the other parameters remain unchanged.  

\section{Experiments and Results}

\subsection{Datasets}

We utilize rumor and stance datasets with propagation structure collected from Twitter\footnote{\url{www.twitter.com}} and Sina Weibo\footnote{\url{www.weibo.com}} as follows. 

\subsubsection{Datasets from Twitter}

To train our model, we refer to three public benchmark datasets with claim veracity labels and propagation tree structure of claims for rumor detection task, namely Twitter15, Twitter16~\cite{ma2018detect}, and PHEME\footnote{This PHEME dataset is not the one used for stance detection~\cite{zubiaga2016analysing}~\cite{derczynski2015pheme}. Instead, it is defined specifically for rumor detection task: \url{https://figshare.com/articles/PHEME_dataset_of_rumours_and_non-rumours/4010619}.}. In each dataset, a group of claims together with their relevant posts are provided, where the claims are treated as bags, and posts are regarded as instances. Each claim is annotated with one of the four class labels: non-rumor, true-rumor, unverified-rumor, and false-rumor. Note that only these bag-level labels are available in the training datasets. We filter out the retweets since they simply repost the claim text without providing any additional information. 

For model evaluation, since both post-level stance and claim-level veracity are needed, we resort to two rumor-stance datasets 
namely RumorEval2019-S~\cite{gorrell2019semeval}\footnote{
We only use the posts from Twitter in RumorEval2019 and discard Reddit data. Note that although the original RumorEval2019-S and SemEval8 datasets contain some same claims, we further extend RumorEval2019-S dataset by including additional 100 non-rumor samples to form a more challenging and general dataset, which includes true rumor, false rumor, unverified rumor, and non-rumor samples. Since SemEval8 only contains true rumor, false rumor, and unverified rumor samples, experiments on these two datasets can evaluate the generalization power of our methods.}
and SemEval8~\citep{derczynski2017semeval, zubiaga2016pheme}. 
The original datasets were used for joint detection of rumors and stances, where each claim is annotated with one of the three veracity classes (i.e., true-rumor, false-rumor and unverified-rumor), and each post was originally assigned one of four stances towards the claim (i.e., agreed, disagreed, appeal-for-more-information, comment). We augmented RumorEval2019-S dataset by gathering additional 100 non-rumor claims together with their relevant posts from Twitter following the method described by~\citet{zubiaga2016analysing}. Then we asked three annotators independent of this work to annotate the stance of posts, \ruichao{the inter-annotator agreement score based on Cohen’s kappa is 0.725}. For stance labels, we convert the original labels into \{S, D, Q, C\} based on a set of rules proposed in~\cite{lukasik2016hawkes}. Table~\ref{tab:TwitterData} and Table~\ref{tab:TwitterRomorStanceFlection} gives the statistics of our Twitter dataset for training and testing. 

\subsubsection{Dataset from Weibo}
We choose a public dataset collected from Weibo\footnote{\url{https://www.dropbox.com/s/46r50ctrfa0ur1o/rumdect.zip?dl=0}} to extend our experimental evaluation. We resort to this dataset for two reasons: (1) We conduct experiments on the Chinese dataset to demonstrate that our methods have a good generalization ability to different languages; (2) Weibo is a two-class rumor detection dataset, and we can demonstrate that our method works well on a different classification scheme and different social media platform. Since it only has claim-level rumor labels, we sample a set of claims to annotate posts' stances for evaluation. Due to the tremendous number of Weibo posts in each claim, we label the first 3,500 posts chronologically for each claim manually. \ruichao{This is to balance coverage and computational feasibility, as well as demonstrate that our model can efficiently verify the rumors being spread in the early stage of their propagation. As a result of post stance annotation, the inter-annotator agreement score based on Cohen's kappa is 0.759}. We provide the statistics of the resulting dataset in Table~\ref{tab:WeiboData}. Note that we follow the method proposed in ~\citet{ma2020attention} to simplify propagation tree structures.

\begin{table*}[t]
  \centering
  \setlength{\abovecaptionskip}{3pt} 
  \setlength{\belowcaptionskip}{0pt}
  \small
  \caption{Statistics of the training (left) and test datasets (right).}
  \resizebox{0.65\textwidth}{!}{
    \begin{tabular}{l|ccc|cc}
    \toprule
     & \multicolumn{3}{c|}{\textbf{Train}} & \multicolumn{2}{c}{\textbf{Test}} \\
    \midrule
    \textbf{Statistics} & \textbf{Twitter15} & \textbf{Twitter16}  & \textbf{PHEME} & \textbf{RumorEval2019-S} & \textbf{SemEval-8} \\
    \midrule
    \# of Rumor categories & 4 & 4 & 4 & 4 & 4 \\
    \# of Stance categories & —— & —— & —— & 4 & 3 \\ 
    \# of claim & 1,308  & 818   & 6,425  & 397 & 297 \\
    \# of Non-rumors & 374 (28.6\%) & 205 (25.1\%) & 4,023 (62.6\%)  & 100 (25.2\%) & —— \\
    \# of False rumors & 370 (28.3\%) & 207 (25.3\%) & 638 (9.9\%)  & 62 (15.6\%) & 62 (20.8\%) \\
    \# of True rumors & 190 (14.5\%) & 205 (25.1\%) & 1,067 (16.6\%) & 137 (34\%) & 137 (46.1\%) \\
    \# of Unverified rumors & 374 (28.6\%) & 201 (24.5\%) & 697 (10.8\%) & 98 (24.7\%) & 98 (33.0\%)\\
    \# posts of Support & ——    & ——    & ——    & 961 (17.8\%) & 645 (15.1\%)\\
    \# posts of Deny & ——    & ——    & ——    & 441 (8.1\%) & 334 (7.8\%)\\
    \# posts of Question & ——    & ——    & ——   & 428 (7.9\%) & 361 (8.5\%) \\
    \# posts of Comment & ——    & ——    & ——    & 3,583 (66.2\%) & 2,923 (68.6\%) \\
    \bottomrule
    \end{tabular}%
    }
    \vspace{-0.3cm}
  \label{tab:TwitterData}%
\end{table*}%

\vspace{-0.3cm}
\begin{table}[htbp]
  \centering
  \small
  \setlength{\abovecaptionskip}{3pt} 
  \setlength{\belowcaptionskip}{0pt}
  \caption{Statistics of the Weibo Dataset.}
  \resizebox{0.4\textwidth}{!}{
    \begin{tabular}{l|ccc}
    \toprule
    \textbf{Statistic} & \textbf{Train} & \textbf{Test}  & \textbf{Total} \\
    \midrule
    \# source tweets & 4,204  & 460   & 4,664 \\
    \# non-rumors & 2,121  & 230   & 2,351 \\
    \# false rumors & 2,083  & 230   & 2,313 \\
    \midrule
    \# posts of Support & -     & 46,697 (22.70\%) & - \\
    \# posts of Deny & -     & 26,122 (12.70\%) & - \\
    \# posts of Question & -     & 22,144 (10.70\%) & - \\
    \# posts of Comment & -     & 110,962 (53.90\%) & - \\
    \midrule
    \# tree nodes & 3,599,731 & 205,925 & 3,805,656 \\
    Avg. posts/tree & 856   & 447   & 816 \\
    Max posts/tree & 59,318 & 3461  & 59,318 \\
    Min posts/tree & 10    & 11    & 10 \\
    \bottomrule
    \end{tabular}%
    }
    \vspace{-0.3cm}
  \label{tab:WeiboData}%
\end{table}%

\begin{table*}[htbp]
  \centering
  \setlength{\abovecaptionskip}{3pt} 
  \setlength{\belowcaptionskip}{0pt}
  \small
  \caption{Statistics of stance distribution for each rumor category in Twitter Dataset.}
  \resizebox{0.8\textwidth}{!}{
    \begin{tabular}{r|cccc|ccc}
    \toprule
  \textbf{\textbf{Dataset}} & \multicolumn{4}{|c|}{\textbf{RumorEval2019-S}} & \multicolumn{3}{|c}{\textbf{SemEval-8}} \\ 
  \midrule    
 \textbf{Statistics} & \textbf{True rumor} & \textbf{False rumor} & \textbf{Unverified rumor} & \textbf{Non-rumor} & \textbf{True rumor} & \textbf{False rumor} & \textbf{Unverified rumor} \\ 
 \midrule 
  \# of Support & 302 (5.58\%) & 102 (1.88\%) & 241 (4.45\%) & 316 (5.84\%) & 302 (7.08\%) & 102 (2.39\%) & 241 (5.65\%)\\
  \# of Deny & 110 (2.03\%) & 88 (1.63\%) & 136 (2.51\%) & 107 (1.98\%) & 110 (2.58\%) & 88 (2.06\%) & 136 (3.19\%)\\
   \# of Question & 156 (2.88\%) & 50 (0.92\%) & 155 (2.86\%) & 67 (1.24\%) & 156 (3.66\%) & 50 (1.17\%) & 155 (3.64\%)\\
   \# of Comment & 1,247 (23.04\%) & 513 (9.48\%) & 1,163 (21.49\%) & 660 (12.19\%) & 1,247 (29.25\%) & 513 (11.45\%) & 1,163 (29.82\%)\\
   \bottomrule
     \end{tabular}}%
     \vspace{-0.3cm}
  \label{tab:TwitterRomorStanceFlection}%
\end{table*}%
\vspace{-0.2cm}

\begin{table}[htbp]
  \centering
  \setlength{\abovecaptionskip}{3pt} 
  \setlength{\belowcaptionskip}{0pt}
  \small
  \caption{Statistics of stance distribution for each rumor category in Weibo Dataset.}
  \resizebox{0.32\textwidth}{!}{
    \begin{tabular}{l|cc}
    \toprule
    \textbf{Statistic} & \textbf{True Rumor} & \textbf{False Rumor} \\
    \midrule
    \# posts of Support & 33,665 (16.3\%) & 13,032 (6.3\%) \\
    \# posts of Deny & 6,256 (3.0\%) & 19,866 (9.7\%) \\
    \# posts of Question & 8,272 (4.0\%) & 13,872 (6.8\%) \\
    \# posts of Comment & 40,966 (19.9\%) & 69,996 (34.0\%) \\
    \bottomrule
    \end{tabular}%
    }
    \vspace{-0.3cm}
  \label{tab:WeiboRomorStanceFlection}%
\end{table}%
\vspace{-0.3cm}

\subsection{Experimental Setup}

We hold out 30\% of the test sets as validation data for tuning hyper-parameters. Due to the imbalanced rumor and stance class distribution, we adopt AUC, micro-averaged and macro-averaged F1 scores, and class-specific F-measure as evaluation metrics in addition to accuracy. 
We implement all the neural models with Pytorch.
When testing on RumorEval2019-S dataset, we train 16 binary classifiers in total considering that there are 4 veracity and 4 stance categories to be determined. And we train 12 binary stance classifiers for testing on SemEval-8 dataset since it contains 3 veracity and 4 stance categories. There are 2 veracity and 4 stance categories in the Weibo dataset, so we develop 8 binary weak classifiers for it.\looseness=-1

\subsection{Stance Detection Performance}

Since our stance detection model is weakly supervised by coarse label (i.e., claim veracity) instead of explicit post-level stance label, we choose to compare with both unsupervised methods and supervised methods:
    (1) \textbf{C-GCN}~\cite{wei2019modeling}: An unsupervised graph convolutional network that classifies the stances by modeling tweets with conversation structure.
    (2) \textbf{Zero-Shot}~\cite{allaway2020zero}: A pre-trained stance detection method that captures relationships between topics, without the need of training examples.
    (3) \textbf{Pre-Rule}~\cite{kobbe2020unsupervised}: An unsupervised method to detect the stance in online debates by identifying effect words based on the pre-defined rules which are designed for support and deny stance. 
    (4) \textbf{BerTweet}~\cite{nguyen2020bertweet}: A language model pre-trained on 850M tweets, which is applied here for post stance detection.
    (5) \textbf{BrLSTM}~\cite{kochkina2017turing}: An LSTM-based model that models the conversational branch structure of posts to detect stance.
    (6) \textbf{MT-GRU}~\cite{ma2018detect}: A multi-task learning model based on RNN 
    to jointly detect rumors and stances by learning the common and task-invariant features.
    (7) \textbf{PE-HCN}~\cite{zhao2020pretrained}: A stance detection method based on the hierarchical capsule network.
    (8) \textbf{JointCL}~\cite{liang2022jointcl}: A zero-shot stance detection model utilizing contrastive learning.
    (9) \textbf{Llama 2-ST}~\cite{touvron2023Llama}: A pre-trained large language model developed by Meta for only stance detection task.
    (10) \textbf{Llama 2-MT}~\cite{touvron2023Llama}: A pre-trained large language model prompted to perform multi-tasks of stance detection and rumor verification together.
    (11) \textbf{ChatGPT-ST}~\cite{NEURIPS2020_1457c0d6}: A pre-trained large language model developed by OpenAI for only stance detection task.
    (12) \textbf{ChatGPT-MT}~\cite{NEURIPS2020_1457c0d6}: A pre-trained large language model prompted to perform multi-tasks of stance detection and rumor verification together.
    (13) \textbf{BU/TD-MIL (\textsc{TrainSet})}~\cite{DBLP:conf/sigir/YangMLG22}: Our MIL-based weakly supervised method with bottom-up/top-down GRU post encoder~\cite{ma2018rumor} for stance detection trained on different \textbf{\textsc{TrainSet}}\footnote{\textbf{\textsc{TrainSet}} can be \textbf{T15}, \textbf{T16} and \textbf{Phe}, which are the short-forms of Twitter15, Twitter16 and PHEME datasets, respectively.}. For the supervised baselines in particular, we use the corresponding validation sets, denoted as \textbf{V}\footnote{\textbf{V} is the corresponding validation sets used to train models from RumorEval2019-S, SemEval-8 or Weibo datasets where models are tested.}, to train them since they need stance labels of posts which do not exist in the \textsc{TrainSet}. To make a fair comparison with these supervised stance detection baselines that are trained on validation sets, we also train our models on the same data although only claim veracity labels are used.
    (14) \textbf{JSDRV~\cite{yang-etal-2024-reinforcement}}: A joint stance detection and rumor verification framework that leverages large language models as the foundation annotators.
    (15) \textbf{BU/TD-MILE (\textsc{TrainSet})}: Our enhanced MIL-based weakly supervised stance detection model trained on different \textsc{TrainSet} with bottom-up/top-down tree transformer~\cite{ma2020debunking} post encoder. 
    (16) \textbf{LLM-MIL (\textsc{TrainSet})}: Our proposed LLM-enhanced MIL framework for stance detection trained on different \textsc{TrainSet}, which allows information flowing between post pairs.

In Table~\ref{tab:StanceResult_Twitter} and Table~\ref{tab:StanceResult_Weibo}, we use the open-source of Zero-Shot and Pre-rule, which does not report AUC. C-GCN, Zero-shot, Pre-Rule, and BerTweet are models without the need of annotated data for stance detection, while BrLSTM, MT-GRU, PE-HCN, and JointCL are four popular supervised stance detection baseline models. Because they need post-level stance labels which is not available in the training set,  we use the post stance annotation in the validation datasets left out from RumorEval2019-S, SemEval-8, and Weibo to train them. To make a comparison with these supervised baselines with the same training data, we train a variant of our methods (i.e., TD/BU-MIL (V), TD/BU-MILE (V), and LLM-MIL (V)) using the same validation dataset by only using the claim veracity labels. 

\begin{table}[t!]
  \centering
  \setlength{\abovecaptionskip}{3pt} 
  \setlength{\belowcaptionskip}{0pt}
  \small
  \caption{Comparison with baselines for stance detection in Twitter Platform. (S: Support; D: Deny; Q: Question; C: Comment)}
  \resizebox{0.86\textwidth}{!}{
    \begin{tabular}{l|ccc|cccc||ccc|cccc}
    \toprule
    \textbf{Dataset} & \multicolumn{7}{c||}{\textbf{RumorEval2019-S}}                 & \multicolumn{6}{c}{\textbf{SemEval8}}                 &  \\
    \midrule
    \multirow{2}[4]{*}{\textbf{Method}} & \multirow{2}[4]{*}{\textbf{AUC}} & \multirow{2}[4]{*}{\textbf{MicF}} & \multirow{2}[4]{*}{\textbf{MacF}} & \textbf{S}     & \textbf{D}     & \textbf{Q}     & \textbf{C}     & \multirow{2}[4]{*}{\textbf{AUC}} & \multirow{2}[4]{*}{\textbf{MicF}} & \multirow{2}[4]{*}{\textbf{MacF}} & \textbf{S}     & \textbf{D}     & \textbf{Q}     & \textbf{C} \\
\cmidrule{5-8}\cmidrule{12-15}          &       &       &       & \textbf{F1}    & \textbf{F1}    & \textbf{F1}    & \textbf{F1}    &       &       &       & \textbf{F1}    & \textbf{F1}    & \textbf{F1}    & \textbf{F1} \\
    \midrule
    C-GCN & 0.633  & 0.629  & 0.416  & 0.331  & 0.173  & 0.429  & 0.730  & 0.610  & 0.625  & 0.411  & 0.327  & 0.161  & 0.430  & 0.728  \\
    Zero-Shot & -     & 0.369  & 0.324  & 0.301  & 0.168  & 0.342  & 0.486  & -     & 0.383  & 0.344  & 0.278  & 0.162  & 0.480  & 0.456  \\
    Pre-Rule & -     & 0.605  & 0.478  & 0.657  & 0.419  & -     & -     & -     & 0.429  & 0.389  & 0.432  & 0.644  & -     & - \\
    BerTweet & 0.643  & 0.619  & 0.492  & 0.497  & 0.203  & 0.513  & 0.753  & 0.621  & 0.611  & 0.428  & 0.512  & 0.131  & 0.326  & 0.742  \\
    \midrule
    BrLSTM (V) & 0.710  & 0.660  & 0.420  & 0.460  & 0.000  & 0.391  & 0.758  & 0.676  & 0.665  & 0.401  & 0.493  & 0.000  & 0.381  & 0.730  \\
    MT-GRU (V) & 0.714  & 0.636  & 0.432  & 0.313  & 0.156  & 0.506  & 0.748  & 0.669  & 0.630  & 0.413  & 0.498  & 0.116  & 0.312  & 0.729  \\
    PE-HCN (V) & 0.714 & 0.638 & 0.499 & 0.401 & 0.374 & 0.471 & 0.751 & 0.670 & 0.635 & 0.493 & 0.410 & \textbf{0.360} & 0.452 & 0.750 \\
    JointCL (V) & 0.716  & 0.639  & 0.505  & 0.532  & 0.210  & 0.516  & 0.760  & 0.678  & 0.640  & 0.475  & 0.536  & 0.136  & 0.478  & 0.751  \\
    \midrule 
    Llama 2-ST	& 0.649 & 0.630 & 0.500 & 0.501 & 0.203 & 0.532 & 0.763 & 0.647 & 0.631 & 0.471 & 0.533 & 0.138 & 0.472 & 0.740 \\
    Llama 2-MT	& 0.647 & 0.632 & 0.500 & 0.502 & 0.199 & 0.533 & 0.766 & 0.649 & 0.630 & 0.473 & 0.534 & 0.142 & 0.471 & 0.742 \\
    ChatGPT-ST	& 0.650 & 0.632 & 0.505 & 0.502 & 0.201 & 0.538 & 0.780 & 0.648 & 0.631 & 0.473 & 0.540 & 0.135 & 0.478 & 0.739 \\
    ChatGPT-MT	& 0.650 & 0.632 & 0.506 & 0.510 & 0.210 & 0.534 & 0.770 & 0.650 & 0.635 & 0.476 & 0.536 & 0.145 & 0.478 & 0.745 \\
    \midrule
    BU-MIL & 0.707 & 0.665 & 0.432 & 0.344 & 0.174 & 0.445 & 0.762 & 0.666 & 0.642 & 0.420 & 0.329 & 0.169 & 0.423 & 0.758 \\
    TD-MIL & 0.722 & 0.691 & 0.434 & 0.344 & 0.179 & 0.467 & 0.767 & 0.669 & 0.651 & 0.426 & 0.335 & 0.175 & 0.430 & 0.763 \\
    JSDRV & 0.725 & 0.723  &  \textbf{0.605}  &  0.546  &  0.476  &  \textbf{0.595}  &  \textbf{0.801}  &  0.725 & 0.705  &  \textbf{0.522}  &  0.563  &  0.216  &  \textbf{0.506}  &  0.801\\
    \midrule
    \midrule
    BU-MILE (V) & 0.719 & 0.644 & 0.522 & 0.536 & 0.255 & 0.516 & 0.780 & 0.690 & 0.666 & 0.475 & 0.534 & 0.130 & 0.479 & 0.756 \\
    BU-MILE (T15) & 0.730 & 0.682 & 0.546 & 0.562 & 0.310 & 0.525 & 0.786 & 0.713 & 0.674 & 0.488 & 0.549 & 0.145 & 0.487 & 0.770 \\
    BU-MILE (T16) & 0.728 & 0.681 & 0.539 & 0.558 & 0.291 & 0.526 & 0.779 & 0.701 & 0.678 & 0.488 & 0.552 & 0.145 & 0.482 & 0.771 \\
    BU-MILE (Phe) & 0.730 & 0.685 & 0.596 & 0.532 & 0.482 & 0.589 & 0.780 & 0.711 & 0.679 & 0.530 & 0.535 & 0.297 & 0.587 &  0.701 \\
    TD-MILE (V) & 0.718 & 0.660 & 0.526 & 0.542 & 0.269 & 0.523 & 0.771 & 0.700 & 0.669 & 0.479 & 0.540 & 0.126 & 0.480 & 0.770 \\
    TD-MILE (T15) & 0.729 & 0.693 & 0.524 & 0.570 & 0.217 & 0.528 & 0.779 & 0.716 & 0.676 & 0.497 & 0.570 & 0.150 & 0.489 & 0.780  \\
    TD-MILE (T16) & 0.729 & 0.692 & 0.542 & 0.560 & 0.300 & 0.526  & 0.780 & 0.700 & 0.679 & 0.492 & 0.556 & 0.149 & 0.485 & 0.778 \\
    TD-MILE (Phe) & 0.731 & 0.694 & 0.543 & 0.554 & 0.298 & 0.536 & 0.783 & 0.720 & 0.680 & 0.506 & 0.539 & 0.209 &  0.497 & 0.780\\
    \midrule
    \textbf{LLM-MIL (V)} & 0.720  & 0.670  & 0.533  & 0.558  & 0.278  & 0.519  & 0.778  & 0.690  & 0.671  & 0.490  & 0.543  & 0.151  & 0.483  & 0.782  \\
    \textbf{LLM-MIL (T15)} & 0.732  & 0.683 & 0.570  & 0.549  & 0.385  & 0.547  & 0.799 & 0.713  & 0.678  & 0.505  & \textbf{0.575} & 0.170  & 0.470 & \textbf{0.805} \\
    \textbf{LLM-MIL (T16)} & 0.735  & 0.698  & 0.570  & 0.540  & 0.383  & 0.558  & 0.799  & 0.702  & 0.697  & 0.508  & 0.567  & 0.189  & 0.480  & 0.796 \\
    \textbf{LLM-MIL (Phe)} & \textbf{0.734 } & \textbf{0.725 } & 0.600 & \textbf{0.551} & \textbf{0.482} & 0.567 & 0.800 & \textbf{0.729} & \textbf{0.708} & 0.512 & 0.565  & 0.218 & 0.464 & 0.801  \\
    \bottomrule
    \end{tabular}%
    }
    \vspace{-0.3cm}
  \label{tab:StanceResult_Twitter}%
\end{table}%

\subsubsection{Twitter Dataset}
In Table~\ref{tab:StanceResult_Twitter}, the first group refers to unsupervised baselines. Zero-Shot and Pre-Rule perform worse than other methods, as they are pre-trained models based on out-of-domain data and cannot generalize well to Twitter data. The results on Q and C by Pre-Rule are absent since its pre-defined linguistic rules are designed for identifying the stance of Support and Deny only. The propagation-based method C-GCN performs better since it attempts to capture structural information in propagation by modeling the neighbors of each post. BerTweet performs best in the first group as it is a BERT-based language model fine-tuned on an enormous Twitter dataset before being applied to downstream tasks.\looseness=-1

The second group considers supervised baselines. MT-GRU performs worse than PE-HCN and JointCL, we speculate that this may be because both PE-HCN and JointCL utilize propagation information. And JointCL performs better among them by leveraging of both context-aware and target-aware features in a contrastive framework, which is beneficial to generalize the stance features to unseen data. BrLSTM improves unsupervised baselines with a large margin in terms of Micro-F1, as BrLSTM is focused on modeling complex propagation structures while MT-GRU is a sequential model. But BrLSTM is poor at classifying denial stance as it is data-driven and the proportion of the denying stance is relatively small in the training data. Our method LLM-MIL (V) outperforms the supervised models clearly in micro-F1 and macro-F1 under the supervision of claim veracity labels only, demonstrating the superiority of our models compared to the fully supervised ones. 

The third group includes results of using LLMs such as Llama 2 and ChatGPT. While Llama 2-ST/MT and ChatGPT-ST/MT outperform all unsupervised baselines, directly using them for stance detection yields unsatisfactory results compared to the supervised model. Furthermore, the performance of the different versions (ST and MT) of Llama 2 and ChatGPT is on the par, suggesting that directly prompting these models to perform both tasks simultaneously yields similar results to performing them separately.
 
The fourth group consists of weakly supervised methods previously proposed by~\citet{DBLP:conf/sigir/YangMLG22} and ~\citet{yang-etal-2024-reinforcement}
. The fifth group contains our enhanced weakly supervised models that leverage bottom-up/top-down tree transformer~\cite{ma2020debunking} as a post encoder. JSDRV performs best because it leverages LLMs as a basic annotator, together with reinforcement learning continuously fine-tuning the model to select high-quality post-stance pairs, thereby mitigating the influence of noise. BU/TD-MILE (V) outperforms BU/TD-MIL (V), and BU/TD-MILE (Phe) achieves superior performance than BU/TD-MIL (Phe) in the stance detection task, as BU/TD-MILE utilizes the tree transformer, which can learn the latent representation of long dependency between words in a post and meaning interaction of posts in a subtree. Moreover, TD-MILE performs comparably with BU-MILE, while the top-down variants appear a little bit better, which indicates that although both benefit from the structure information utilizing tree transformer, the bottom-up model may suffer from larger information loss. This result is consistent with the findings presented in~\cite{ma2018rumor,ma2020debunking}.

Our LLM-enhanced MIL method achieves the best Micro-F1 score when trained on the PHEME dataset, which has significantly exceeded all methods when training data is large enough. It suggests that the LLM-enhanced MIL framework is not limited by propagation direction and can make better use of the propagation structure and LLMs than any other baseline methods. We also observe that JSDRV achieves the best Macro-F1 score, followed by LLM-MIL(Phe). The class-specific F1 score improves for the "Support" and "Deny" categories, but decreases for the "Question" category substantially. This might be because questioning posts often contain inherent doubts and uncertainties, so LLMs might struggle to capture the key clues, leading to hallucinations and generating unreliable explanations.
In particular, BU-MILE, TD-MILE, and LLM-MIL (Phe) perform better than their counterparts trained on Twitter15/16 datasets. This might be attributed to PHEME containing more claims than Twitter15/16 for weak supervision. We conjecture that the performance of our method can be further improved when trained on larger datasets. 

\subsubsection{Weibo Dataset}

The stances in the Weibo dataset are labeled with four-way classes (Support, Deny, Comment, and Question), and Table~\ref{tab:StanceResult_Weibo} shows the stance detection result across all the compared methods. At first glance, Weibo data follows a similar performance to Twitter data. However, there are still significant differences worth discussing.

Among unsupervised baselines, BerTweet performs worse than C-GCN and Pre-Rule, as BerTweet is fine-tuned with the Chinese language whereas it is pre-trained on the English language corpus. Unsurprisingly, it may not perform as well as it should. However, BerTweet performs better than Zero-Shot, indicating that a large amount of training corpus still has a strong inspiring effect on the unknown post representation learning process. The second and third group shows a similar performance compared with the Twitter dataset, which is also a four-category classification problem on social media with regard to user opinions.\looseness=-1

The fourth group consists of our weakly supervised method BU/TD-MIL proposed in ~\citet{DBLP:conf/sigir/YangMLG22} and ~\citet{yang-etal-2024-reinforcement}, and the fifth group refers to our enhanced weakly supervised methods with bottom-up/top-down tree transformer~\cite{ma2020debunking} post encoder. It is worth mentioning that TD-MILE (V) outperforms JonintCL (V) on the Weibo dataset, indicating that our MIL-based method has the potential to surpass supervised models. This might be because the prototype generation process in JointCL is designed for the English corpus and the generalization to the Chinese corpus is limited. 

Our proposed LLM-enhanced MIL models still outperform all the baselines on Micro-F1 score, and only underperform JSDRV on Macro-F1 score, which is consistent with the results we have obtained on the three Twitter datasets. We conjecture that our models enhance the representation learning of content information while the important backbone structural information can also be retained. 
\begin{table}[t!]
  \centering
  \setlength{\abovecaptionskip}{3pt} 
  \setlength{\belowcaptionskip}{0pt}
  \caption{Comparison with baselines for stance detection on Weibo Dataset. (S: Support; D: Deny; Q: Question; C: Comment)}
  \small
  \resizebox{0.48\textwidth}{!}{
    \begin{tabular}{l|ccc|cccc}
    \toprule
    \multirow{2}[4]{*}{\textbf{Method}} & \multirow{2}[4]{*}{\textbf{AUC}} & \multirow{2}[4]{*}{\textbf{MicF}} & \multirow{2}[4]{*}{\textbf{MacF}} & \textbf{S}     & \textbf{D}     & \textbf{Q}     & \textbf{C} \\
\cmidrule{5-8}          &       &       &       & \textbf{F1}    & \textbf{F1}    & \textbf{F1}    & \textbf{F1} \\
    \midrule
    C-GCN & 0.630  & 0.631  & 0.417  & 0.329  & 0.172  & 0.433  & 0.732  \\
    Zero-Shot & -     & 0.371  & 0.325  & 0.311  & 0.169  & 0.343  & 0.476  \\
    Pre-Rule & -     & 0.615  & 0.536  & 0.651  & 0.421  & -     & - \\
    BerTweet & 0.625  & 0.601  & 0.405  & 0.339 & 0.140  & 0.402  & 0.738  \\
    \midrule
    BrLSTM (V) & 0.709  & 0.650  & 0.404  & 0.462  & 0.000  & 0.392  & 0.763  \\
    MT-GRU (V) & 0.716  & 0.634  & 0.433  & 0.316  & 0.155  & 0.509  & 0.752  \\
    PE-HCN (V) & 0.715 & 0.639 & 0.464 & 0.435 & 0.201 & 0.465 & 0.754\\
    JointCL (V) & 0.719  & 0.652  & 0.492  & 0.499  & 0.211  & 0.500  & 0.756  \\
    \midrule
    Llama 2-ST & 0.713 & 0.629 & 0.424 & 0.395  & 0.161  & 0.373  & 0.765\\
    Llama 2-MT & 0.712 & 0.629 & 0.424 & 0.391 & 0.160 & 0.375 & 0.763\\
    ChatGPT-ST & 0.715 & 0.631 & 0.425 & 0.397 & 0.167 & 0.370 & 0.767\\
    ChatGPT-MT & 0.716 & 0.632 & 0.424 & 0.389 & 0.170 & 0.374 & 0.764\\
    \midrule
    BU-MIL & 0.719 & 0.650 & 0.532 & 0.518 & 0.360 & 0.505 & 0.748 \\
    TD-MIL & 0.719 & 0.653 & 0.549 & 0.520 & 0.365 & 0.569 & 0.743 \\ 
    JSDRV & 0.721 & 0.690 & \textbf{0.578} & 0.548 & 0.385 & \textbf{0.612} & 0.766\\
    \midrule
    \midrule
    BU-MILE (V) & 0.719 & 0.651 & 0.535 & 0.529 & 0.371 & 0.520 & 0.720 \\
    BU-MILE & 0.722 & 0.680 & 0.573 & 0.544 & 0.380 & 0.606 & 0.761\\
    TD-MILE (V) & 0.720 & 0.655 & 0.537 & 0.530 & 0.375 & 0.520 & 0.723\\
    TD-MILE & 0.721 & 0.683 & 0.574 & 0.544 & 0.382 & 0.609 & 0.761\\
    \midrule
    \textbf{LLM-MIL (V)} & 0.720 & 0.663  & 0.540  & 0.539  & 0.375  & 0.519  & 0.726  \\
    \textbf{LLM-MIL}  & \textbf{0.722} & \textbf{0.699} & 0.570 & \textbf{0.549} & \textbf{0.389} & 0.574 & \textbf{0.768} \\
    \bottomrule
    \end{tabular}%
    }
    \vspace{-0.3cm}
  \label{tab:StanceResult_Weibo}%
\end{table}%

\begin{table}[t!]
  \centering
  \setlength{\abovecaptionskip}{3pt} 
  \setlength{\belowcaptionskip}{0pt}
  \small
  \caption{Results on Rumor Detection in Twitter Platform. (N: Non-rumor; F: False rumor; T: True rumor; U: Unverified rumor)}
  \resizebox{0.86\textwidth}{!}{
    \begin{tabular}{l|ccc|cccc||ccc|ccc}
    \toprule
    \textbf{Dataset} & \multicolumn{7}{c||}{\textbf{RumorEval2019-S}}                 & \multicolumn{6}{c}{\textbf{SemEval8}} \\
    \midrule
    \multirow{2}[4]{*}{\textbf{Method}} & \multirow{2}[4]{*}{\textbf{AUC}} & \multirow{2}[4]{*}{\textbf{MicF}} & \multirow{2}[4]{*}{\textbf{MacF}} & \textbf{T}     & \textbf{F}     & \textbf{U}     & \textbf{N}     & \multirow{2}[4]{*}{\textbf{AUC}} & \multirow{2}[4]{*}{\textbf{MicF}} & \multirow{2}[4]{*}{\textbf{MacF}} & \textbf{T}     & \textbf{F}     & \textbf{U} \\
\cmidrule{5-8}\cmidrule{12-14}          &       &       &       & \textbf{F1}    & \textbf{F1}    & \textbf{F1}    & \textbf{F1}    &       &       &       & \textbf{F1}    & \textbf{F1}    & \textbf{F1} \\
    \midrule
    AE & 0.601 & 0.513 & 0.487 & 0.562 & 0.307 & 0.479 & 0.601 & 0.569 & 0.502 & 0.439 & 0.403 & 0.398 & 0.517 \\
    UFD & 0.672 & 0.567 & 0.549 & 0.568 & 0.483 & 0.506 & 0.641 & 0.491 & 0.505 & 0.507 & 0.500 & 0.439 & 0.583 \\
    BerTweet & 0.693  & 0.752  & 0.729  & 0.755  & 0.647  & 0.699  & 0.813  & 0.684  & 0.660  & 0.697  & 0.708  & 0.624  & 0.756  \\ 
    ptVAE & 0.690 & 0.748 & 0.730 & 0.762 & 0.658 & 0.690 & 0.810 & 0.685 & 0.657 & 0.697 & 0.700 & 0.633 & 0.758 \\
    \midrule
    GCAN  & 0.693  & 0.645  & 0.253  & 0.249  & 0.310  & 0.113  & 0.339  & 0.688  & 0.645  & 0.255  & 0.241  & 0.326  & 0.198  \\
    TD-RvNN & 0.881  & 0.753  & 0.727  & 0.755  & 0.666  & 0.673  & 0.815  & 0.885  & 0.748  & 0.694  & 0.712  & 0.617  & 0.753  \\
    PLAN  & 0.892  & 0.800  & 0.792  & 0.819  & 0.760  & 0.780  & 0.812  & 0.880  & 0.794  & 0.753  & 0.741  & 0.694  & 0.825  \\
    DDGCN & 0.897  & 0.806  & 0.795  & 0.835  & 0.763  & 0.766  & 0.817  & 0.890  & 0.796  & 0.753  & 0.741  & 0.694  & 0.825  \\
    \midrule
    Llama 2-ST	& 0.682 & 0.754 & 0.450 & 0.660 & 0.271 & 0.400 & 0.469 & 0.682 & 0.746 & 0.424 & 0.632 & 0.260 & 0.380 \\
    Llama 2-MT	& 0.685 & 0.758 & 0.465 & 0.678 & 0.301 & 0.403 & 0.478 & 0.682 & 0.756 & 0.427 & 0.635 & 0.263 & 0.382 \\
    ChatGPT-ST	& 0.682 & 0.756 & 0.465 & 0.678 & 0.301 & 0.403 & 0.478 & 0.680 & 0.758 & 0.430 & 0.638 & 0.263 & 0.390 \\
    ChatGPT-MT	& 0.686 & 0.762 & 0.470 & 0.680 & 0.311 & 0.409 & 0.480 & 0.683 & 0.758 & 0.430 & 0.638 & 0.268 & 0.384	\\
    \midrule
    BU-MIL & 0.904 & 0.776 & 0.763 & 0.793 & 0.666 & 0.770 & 0.833 & 0.902 & 0.763 & 0.729 & 0.728 & 0.649 & 0.809\\
    TD-MIL & 0.917 & 0.809 & 0.776 & 0.826 & 0.659 & 0.669 & 0.852 & 0.908 & 0.798 & 0.741 & 0.741 & 0.672 & 0.810 \\
    JSDRV & 0.908 & 0.842  &  \textbf{0.804}  & 0.829  &  0.774  &  \textbf{0.824}  &  0.787  &  0.909 & 0.834  &  \textbf{0.784}  &  0.820  &  0.741  &  0.792\\
    \midrule
    \midrule
    BU-MILE (T15) & 0.906 & 0.815 & 0.782 & 0.826 & 0.758 & 0.717 & 0.827 & 0.910 & 0.808 & 0.759 & 0.756 & 0.693 &  0.828\\
    BU-MILE (T16) & 0.901 & 0.810 & 0.786 & 0.825 & 0.762 & 0.719 & 0.838 & 0.899 & 0.800 & 0.755 & 0.759 & 0.692 & 0.814 \\
    BU-MILE (Phe) & 0.915 & 0.816 & 0.785 & 0.825 & 0.770 & 0.715 & 0.830 & 0.910 & 0.810 & 0.758 & 0.763 & 0.698 & 0.813 \\
    TD-MILE (T15) & 0.908 & 0.816 & 0.786 & 0.829 & 0.760 & 0.720 & 0.835 & 0.910 & 0.808 & 0.764 & 0.776 & 0.700 & 0.816 \\
    TD-MILE (T16) & 0.909 & 0.813 & 0.789 & 0.828 & 0.760 & 0.729 & 0.839 & 0.898 & 0.803 & 0.769 & 0.771 & 0.709 & 0.827 \\
    TD-MILE (Phe) & 0.901 & 0.820 & 0.789 & 0.829 & 0.761 & 0.738 & \textbf{0.828} & 0.910 & 0.811 & 0.775 & 0.778 & 0.711 & \textbf{0.836} \\
    \midrule
    \textbf{LLM-MIL (T15)} & 0.908  & 0.846  & 0.783  & 0.830 & 0.778  & 0.802  & 0.722  & 0.908  & 0.831  & 0.780  & 0.827  & \textbf{0.749}  & 0.767  \\
    \textbf{LLM-MIL (T16)} & 0.901  & 0.846  & 0.789  & 0.839  & 0.774  & 0.806  & 0.737  & 0.900  & 0.835  & 0.781  & 0.829  & \textbf{0.749}  & 0.765 \\
    \textbf{LLM-MIL (Phe)} & \textbf{0.920 } & \textbf{0.848 } & 0.802  & \textbf{0.840}  & \textbf{0.779}  & 0.812  & 0.777 & \textbf{0.912 } & \textbf{0.836 } & 0.781  & \textbf{0.831 } & 0.742  & 0.771 \\
    \bottomrule
    \end{tabular}%
    }
    \vspace{-0.3cm}
  \label{tab:RumorResult_Twitter}%
\end{table}%

\begin{table}[t!]
  \centering
  \setlength{\abovecaptionskip}{3pt} 
  \setlength{\belowcaptionskip}{0pt}
  \small
  \caption{Rumor Detection Results on Weibo Dataset.}
  \resizebox{0.55\textwidth}{!}{
    \begin{tabular}{l|ccc|ccc|ccc}
    \toprule
    \multirow{2}[4]{*}{\textbf{Method}} & \multirow{2}[4]{*}{\textbf{AUC}} & \multirow{2}[4]{*}{\textbf{MicF}} & \multirow{2}[4]{*}{\textbf{MacF}} & \multicolumn{3}{c|}{\textbf{Rumor}} & \multicolumn{3}{c}{\textbf{Non-Rumor}} \\
\cmidrule{5-10}          &       &       &       & \textbf{Pre.}  & \textbf{Rec.}  & \textbf{F1}    & \textbf{Pre.}  & \textbf{Rec.}  & \textbf{F1} \\
    \midrule
    AE & 0.625 & 0.620 & 0.594 & 0.631 & 0.567 & 0.597 & 0.572 & 0.610 & 0.590 \\
    UFD & 0.737 & 0.635 & 0.624 & 0.645 & 0.609 & 0.626 & 0.604 & 0.639 & 0.621  \\
    BerTweet & 0.806  & 0.854 & 0.728 & 0.751 & 0.713 & 0.732 & 0.700 & 0.752 & 0.725 \\
    ptVAE & 0.805 & 0.856 & 0.732 & 0.76  & 0.71  & 0.734 & 0.703 & 0.759 & 0.730\\
    \midrule
    GCAN  & 0.903  & 0.848  & 0.738 & 0.738 & 0.73  & 0.734 & 0.719 & 0.766 & 0.742 \\
    TD-RvNN & 0.968  & 0.946  & 0.938 & 0.942 & 0.95  & 0.946 & 0.910 & 0.953 & 0.931  \\
    PLAN  & 0.962  & 0.948  & 0.935 & 0.942 & 0.938 & 0.940 & 0.910 & 0.950 & 0.930 \\
    GAN & -	& 0.952  & 0.943 & 0.940 & 0.956  & 0.948 & \textbf{0.952} & 0.923 & \textbf{0.937} \\
    DDGCN &  0.960  & 0.940  & 0.934 & 0.945 & 0.938 & 0.941 & 0.914 & 0.938 & 0.926\\
    \midrule
    Llama 2-ST & 0.814 & 0.860 & 0.734 & 0.756 & 0.731 & 0.743 & 0.714 & 0.736 & 0.725\\
    Llama 2-MT & 0.814 & 0.861 & 0.735 & 0.758 & 0.729 & 0.743 & 0.715 & 0.740 & 0.727\\
    ChatGPT-ST & 0.813 & 0.865 & 0.738 & 0.763 & 0.729 & 0.746 & 0.718 & 0.742 & 0.730\\
    ChatGPT-MT & 0.815 & 0.866 & 0.739 & 0.76  & 0.732 & 0.746 & 0.721 & 0.745 & 0.733\\
    \midrule
    BU-MIL & 0.893 & 0.948 & 0.932 & 0.959 & 0.925 & 0.942 & 0.910 & 0.933 & 0.921 \\
    TD-MIL & 0.895 & 0.949 & 0.935 & 0.953 & 0.941 & 0.947 & 0.912 & 0.935 & 0.923 \\
    JSDRV & 0.968 & 0.960 & \textbf{0.945} & \textbf{0.961} & 0.955 & 0.958 & 0.920 & \textbf{0.944} & 0.932 \\
    \midrule
    \midrule
    BU-MILE & \textbf{0.972} & 0.958 & 0.939 & 0.96  & 0.948 & 0.954 & 0.914 & 0.936 & 0.925 \\
    TD-MILE & 0.970 & 0.959  & 0.940 & 0.958 & 0.949 & 0.953 & 0.917 & 0.938 & 0.927 \\
    \midrule
    \textbf{LLM-MIL} & \textbf{0.972} & \textbf{0.963} & 0.944 & \textbf{0.961} & \textbf{0.957} & \textbf{0.959} & 0.919 & 0.940 & 0.929 \\
    \bottomrule
    \end{tabular}%
    }
    \vspace{-0.3cm}
  \label{tab:RumorResult_Weibo}%
\end{table}%

\subsection{Rumor Detection Performance}

We compare our methods with the following state-of-the-art structured and unstructured rumor detection baselines: 
    (1) \textbf{AE}~\cite{zhang2017detecting}: An unsupervised method based on autoencoder to perform rumor detection. 
    (2) \textbf{UFD}~\cite{yang2019unsupervised}: An unsupervised rumor detection method based on user opinion and user credibility.
    (3) \textbf{BerTweet}~\cite{nguyen2020bertweet}: A language model pre-trained on 850M tweets, which is applied here for rumor detection.
    (4) \textbf{ptVAE}~\cite{fang2023unsupervised}: A tree variational autoencoder model for unsupervised rumor detection leveraging tree propagation structure.
    (5) \textbf{GCAN}~\cite{lu2020gcan}: A graph-aware co-attention model utilizing retweet structure to verify the source tweet.
    (6) \textbf{TD-RvNN}~\cite{ma2020attention}: A tree-structured attention network for rumor detection with top-down propagation structure.
    (7) \textbf{PLAN}~\cite{khoo2020interpretable}: A transformer based rumor detection model utilizing user interactions.
    (8) \textbf{GAN}~\cite{ma2021improving}: A transformer-based detection model leveraging generative adversarial learning\footnote{Generative adversarial learning is usually applied in two category samples. So, we only report experimental results on the Weibo dataset. And the original code does not report the AUC score.}.
    (9) \textbf{DDGCN}~\cite{sun2022ddgcn}: A rumor detection utilizing the dynamics of propagation messages and external background knowledge.
    (10) \textbf{Llama 2-ST/MT}~\cite{touvron2023Llama}: A large language model pre-trained on various domains myriad corpus, we use it for single rumor verification task, and joint stance detection and rumor verification task, respectively.
    (11) \textbf{ChatGPT-ST/MT}~\cite{NEURIPS2020_1457c0d6}: A large language model developed by OpenAI, we use it for single rumor verification task, and joint detection of stance and rumor veracity, respectively.
    (12) \textbf{BU/TD-MIL(\textsc{TrainSet})}~\cite{DBLP:conf/sigir/YangMLG22}: Our MIL-based weakly supervised method with bottom-up/top-down GRU post encoder~\cite{ma2018rumor} for rumor detection trained on different \textsc{TrainSet}.
    (13) \textbf{JSDRV}~\cite{yang-etal-2024-reinforcement}:  A joint stance detection (SD) and rumor verification framework that leverages large language models as the foundation annotators.
    (14) \textbf{BU/TD-MILE(\textsc{TrainSet})}: Our enhanced MIL-based weakly supervised rumor detection method trained on different \textsc{TrainSet} utilizing bottom-up/top-down tree transformer~\cite{ma2020debunking} as post encoder.
    (15) \textbf{LLM-MIL(\textsc{TrainSet})}: Our newly proposed MIL-based model with graph transformer trained on different \textsc{TrainSet} for rumor detection.\looseness=-1

\subsubsection{Twitter Dataset}
In Table~\ref{tab:RumorResult_Twitter}, the first group relates to unsupervised baselines. AE and UFD perform worse than ptVAE, as AE and UFD merely rely on content and user information while ptVAE additionally utilizes propagation structure with a variational autoencoder. BerTweet outperforms ptVAE slightly, considering it is a large language model trained on an enormous tweets corpus and has better generalizability to unseen data.

The second group consists of supervised baselines. We observe that GCAN is worse than the other systems, which only consider local structures, such as direct neighborhoods. In contrast, TD-RvNN models the global propagation contexts by aggregating the entire propagation information recursively. PLAN outperforms TD-RvNN as it incorporates post-level cross-attention, taking into account the interaction of any pair of posts, which has superior representation ability than RvNN. DDGCN performs best across all supervised baselines given that it not only incorporates the dynamics of message propagation feature but also considers the dynamics of background knowledge in a unified framework, which provides additional evidential information. 

The third group includes LLMs like Llama 2 and ChatGPT. While Llama 2-ST/MT and ChatGPT-ST/MT outperform all unsupervised baselines, they perform worse in rumor detection compared to the supervised model, indicating that directly using LLMs to predict rumors may suffer from inherent issues of LLMs.

The fourth group refers to weakly supervised methods proposed in~\citet{DBLP:conf/sigir/YangMLG22} and ~\citet{yang-etal-2024-reinforcement}, and the fifth group refers to our enhanced weakly supervised methods that replace GRU post encoder with tree transformer, as aforementioned. We can see that BU/TD-MILE surpasses BU/TD-MIL by a large margin in the rumor detection task across all training datasets. This can be attributed to the capability of BU/TD-MILE that not only utilizes propagation structure but also leverages the tree transformer to model the long dependency between words as well as posts, which further enhances the post representation. Furthermore, TD-MILE outperforms BU-MILE, as the former considers both local and global contexts while the latter only considers the local contexts during the stance aggregation process. We also observe that our enhanced weakly supervised methods outperform all the unsupervised and supervised baselines on all the training sets, demonstrating the strength of representation enhanced by the tree transformer.\looseness=-1

Our LLM-enhanced MIL method achieves the best Micro-F1 score when trained on the PHEME dataset, which has significantly exceeded all methods when training data is large enough. It suggests that the LLM-enhanced MIL framework is not limited by propagation direction and can make better use of the propagation structure and LLMs than any other baseline methods. We observe that JSDRV achieves the best Macro-F1 score, followed by LLM-MIL(Phe). Similary, the class-specific F1 score gets improved for the "Support" and "Deny" categories while dropping for the "Question" category substantially due to the similar reason of hallucinations and generation of unreliable explanations.
In particular, BU-MILE, TD-MILE, and LLM-MIL (Phe) perform better than their counterparts trained on Twitter15/16 datasets. This might be attributed to PHEME containing more claims than Twitter15/16 for weak supervision. We conjecture that the performance of our method can be further improved when trained on larger datasets. 

Our LLM-enhanced MIL method achieves better Micro- and Macro-F1 scores compared with TD-MILE (Phe) on the test sets. This can be explained by our design that LLM-MIL not only uses graph attention and explanations generated by ChatGPT for post representation but also aggregates stances with the graph-based attention mechanism, which can capture complex interactions between related posts under each claim and promote the impact of indicative stances on the rumor detection task. In contrast, TD-MILE only utilizes local subtree contexts for post representation. Meanwhile, the PHEME dataset has more claims and relevant posts than the Twitter15 and Twitter16 datasets, making the relationship between posts naturally more complex. Then graph attention and ChatGPT can take advantage of the complex graph structure. We also observe that our LLM-enhanced method achieves the best Micro-F1 score when trained on the PHEME dataset, while JSDRV achieves the best Macro-F1 score, followed by LLM-MIL (Phe). The class-specific F1 score is improved for "True Rumor" and "False Rumor" categories, but worsens for the "Unverified Rumor" and "Non Rumor" categories, implying hallucination of ChatGPT with questioning posts, which could harm the model's ability to identify unverified rumors and non-rumors.

\subsubsection{Weibo Dataset}

The claims in the Weibo dataset are labeled with two categories (e.g., rumor, and non-rumor). 
While a similar trend of performance on Weibo dataset is observed in Table~\ref{tab:RumorResult_Twitter}, there are some salient differences compared with multi-category classification results.\looseness=-1

Different from the multi-class dataset on Twitter,  ptVAE outperforms BerTweet in the first group. This might suggest that binary classification is less challenging than finer-grained classification and the propagation-based autoencoder can introduce extra structure for post representation learning. Meanwhile, BerTweet, which is pre-trained on English datasets, has not been fine-tuned on the Chinese Weibo dataset. Therefore, the performance of BerTweet can be somewhat limited.\looseness=-1

In the second group relating to supervised methods, PLAN outperforms DDGCN. This can be explained by the intuition that the transformer-based PLAN model can enhance the representation of different features of samples under the long dependency posts structure, while the external knowledge utilized in DDGCN may scatter the attention of the model given a large undirected graph. It is worth noting that GAN performs best among all the supervised baselines. This implies that generative adversarial learning can improve the feature learning process of samples in two opposing types, which shows superior classification capabilities on unseen data.

The third group shows a similar trend compared with the Twitter dataset. And in the fourth group, BU-MIL underperforms PLAN, GAN, and DDGCN and achieves comparable results on MicF with TD-RvNN. We speculate that PLAN and GAN both utilize a transformer model that has a strong ability in post representation, while BU-MIL just uses a GRU encoder. 
However, in the fifth group, BU/TD-MILE outperforms all the baselines as BU/TD-MILE can fully utilize both stance features and structure information explicitly with the graph structure. 

Finally, our LLM-MIL performs only worse than JSDRV with the decreased class-specific F1 score of "Non Rumor" class. However, it performs best compared with all other models on Micro-F1 score, thanks to its improved post representation with ChatGPT and the hierarchical stance graph attention mechanism that can capture complex post interaction and rumor indicative features.

\subsection{Ablation Study}
To evaluate the impact of each component, we perform an ablation study based on the best-performed LLM-MIL (Phe) models on RumorEval2019-S dataset by removing some components: 
(1) \textbf{w/o s}: remove po\underline{s}t-level representation that realized in Equation~\ref{equ:PostLevelRepre};
(2) \textbf{w/o e}: remove \underline{e}xplanation-level attention that realized in Equations~\ref{equ:ExplanationEncode};
(3) \textbf{w/o p}: replace the \underline{p}ost encoder (Equations~\ref{equ:Token-levelRepre}-\ref{equ:ExplanationEncode}) with GRU; 
(4) \textbf{w/o o}: remove the l\underline{o}cal attention mechanism (realized in Equation~\ref{equ:GraphLocalAtt}) in stance aggregation process;
(5) \textbf{w/o g}: replace the \underline{g}lobal attention mechanism (realized in Equation~\ref{equ:GraphRumorPredict}) with mean operator for stance aggregation;
(6) \textbf{w/o h}: replace the \underline{h}ierarchical attention mechanism (Equation[\ref{equ:GraphLocalAtt}-\ref{equ:GraphRumorPredict}]) with general dot product attention~\cite{vaswani2017attention} for stance aggregation;
(7) \textbf{w/o a}: replace the explanation guided \underline{a}ttention mechanism (Equations~\ref{equ:betaattention}-\ref{equ:finalclaimpredict}) in binary classifiers aggregation process with mean operator.


As shown in Table~\ref{tab:ablationstudy}, every ablation leads to a deterioration of the results, which demonstrates these components' importance. \textbf{w/o s} implies that post-level representation can help stance representation. \textbf{w/o e} indicates that the explanation-level representation generated by LLM is useful to our method. And \textbf{w/o p} indicates that replacing token-level and post-level representation with GRU further reduces the performance of our model, demonstrating the effectiveness of the post-level representation. \textbf{w/o o} entails that the local attention mechanism improves the stance aggregation process in the model. \textbf{w/o g} exhibits that the global attention mechanism is more vital than the local attention mechanism to stance aggregation. \textbf{w/o h} reveals that our claim-guided hierarchical attention mechanism can effectively combine the local and global attention from the two different levels. \textbf{w/o a} gets the lowest performance scores, where AUC, MicF, and MacF decrease about 8\%, 21\%, and 27\% for rumor detection and 12\%, 24\%, and 22\% for stance detection, which suggests that binary classifiers aggregation based on weighted collective attention mechanism is the most important component in our method.\looseness=-1 

\begin{table}[t!]
  \centering
  \setlength{\abovecaptionskip}{3pt} 
  \setlength{\belowcaptionskip}{0pt}
  \small
  \caption{Ablation study results.}
  \resizebox{0.39\textwidth}{!}{
    \begin{tabular}{l|rrr||rrr}
    \toprule
          & \multicolumn{3}{c||}{\textbf{Rumor Result}} & \multicolumn{3}{c}{\textbf{Stance Result}} \\
    \midrule
    \textbf{Method} & \multicolumn{1}{l}{\textbf{AUC}} & \multicolumn{1}{l}{\textbf{MicF}} & \multicolumn{1}{l||}{\textbf{MacF}} & \multicolumn{1}{l}{\textbf{AUC}} & \multicolumn{1}{l}{\textbf{MicF}} & \multicolumn{1}{l}{\textbf{MacF}} \\
    \midrule
    w/o s &  0.918 & 0.832 & 0.800 & 0.734 & 0.693 & 0.566\\
    w/o e &  0.917 & 0.819 & 0.798 & 0.732 & 0.692 & 0.562\\
    w/o p &  0.914 & 0.821 & 0.783 & 0.727 & 0.674 & 0.553\\
    w/o o &  0.915 & 0.825 & 0.795 & 0.729 & 0.686 & 0.560\\
    w/o g & 0.915 & 0.822 & 0.795 & 0.726 & 0.685 & 0.555\\
    w/o h & 0.911 & 0.815 & 0.776 & 0.725 & 0.673 & 0.551\\
    w/o a & 0.912 & 0.814 & 0.771 & 0.722 & 0.672 & 0.548 \\
    \midrule
    \textbf{LLM-MIL} & 0.920 & 0.848 & 0.802 & 0.734 & 0.725 & 0.600 \\
    \bottomrule
    \end{tabular}%
    }
  \label{tab:ablationstudy}%
\end{table}%

\subsection{Sensitivity Study}
\begin{figure*}[t!]
\centering
\setlength{\abovecaptionskip}{3pt} 
\setlength{\belowcaptionskip}{0pt}
\includegraphics[width=3.4in]{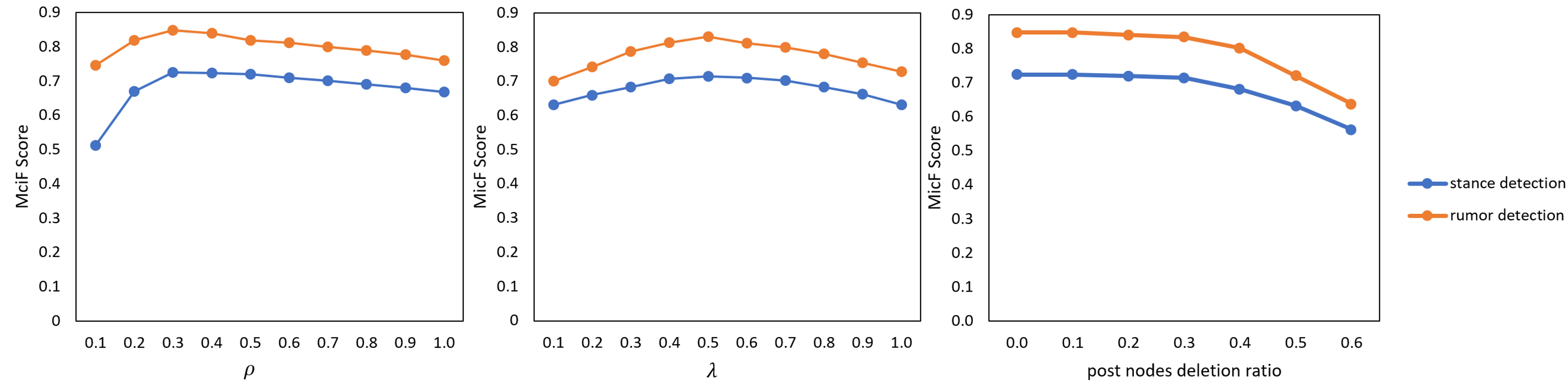}
\caption{Impact of the retention ratio $\rho$,  $\lambda$, \ruichao{and the ratio of post node deletion for rumor and stance detection tasks}.}
\label{fig:sensitivity}
\vspace{-0.5cm}
\end{figure*}

We conduct a sensitivity study to see the impact of the hyper-parameter retention ratio $\rho$ for post-level representation and $\lambda$ in local attention for binary rumor classification. \ruichao{Additionally, we also conduct a study to see the impact of randomly deleting posts nodes for final experiment results.} Using LLM-MIL (Pheme) model, we show the variation of micF score on stance and rumor detection tasks with different $\rho$, $\lambda$, \ruichao{and the ratio of post nodes deletion. We present the performance of (a) various values of $\rho$ with $\lambda = 0.5$, (2) different values of $\lambda$ with $\rho = 0.5$, and (c) different ratios of post nodes deletion with $\rho=0.3$ and $\lambda = 0.5$.}
Figure~\ref{fig:sensitivity} indicates that (1) when the retention ration $\rho$ = 0.3, the performance of LLM-MIL(Pheme) reaches its peak; (2) when retention ration $\lambda$ = 0.5, the performance of LLM-MIL(Pheme) reaches its peak; \ruichao{and (3) when the ratio of post nodes deletion reaches 0.3, the performance of LLM-MIL(Pheme) begins to decline, suggesting that our model is basically resilient to incomplete propagation structure and noise.}


\subsection{Case Study}

To get an intuitive understanding of the LLM-MIL models' weak binary classifiers and final aggregation model, we design experiments to analyze the behavior of LLM-MIL (Phe) to gain insights into its superior performance in the group of LLM-MIL models.\looseness=-1

\subsubsection{Case Study for Weak Binary Classifiers}

To get an intuitive understanding of the \textit{Hierarchical Stance Tree Attention Mechanism}, we sample two graphs from RumorEval2019-S test set that the source claims have been correctly classified as True and False rumor, and show their corresponding local and stance explanation-guided global attention.\looseness=-1

\begin{figure}[t!]
\centering
\setlength{\abovecaptionskip}{3pt} 
\setlength{\belowcaptionskip}{0pt}
\subfigure[True Rumor]{
\includegraphics[scale=0.4]{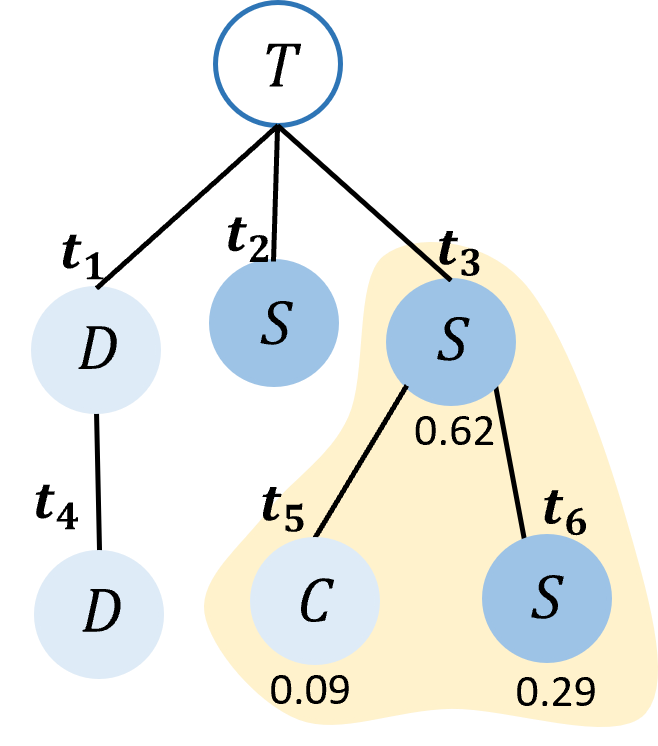}
\label{fig:CaseStudyTrue}}
\subfigure[False rumor]{
\includegraphics[scale=0.4]{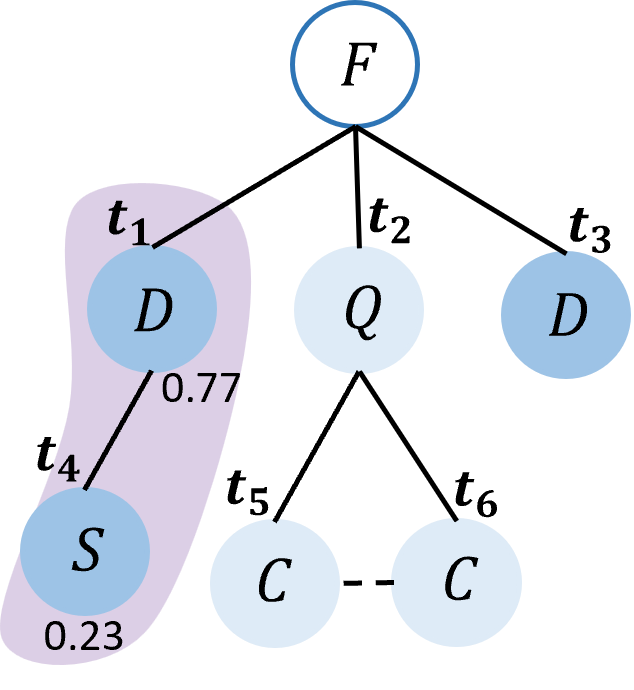}
\label{fig:CaseStudyFalse}}
\caption{Case Study for Claim-guided Hierarchical Stance Tree Attention.}\label{fig:CaseStudy_Local&GlobalAtt}
\vspace{-0.3cm}
\end{figure}

For local attention, we only show one node and its neighbors' averaged local attention scores over all the binary classifiers for the sake of conciseness: We select node $t_3$ in the true rumor case, and node $t_1$ in the false rumor case as displayed in Figure~\ref{fig:CaseStudy_Local&GlobalAtt}. As for global attention, we firstly compute the nodes' averaged global attention scores over all binary classifiers, and then mark the top-3 important stances with solid blue ovals in the true and false rumor cases, respectively. \looseness=-1

We have the following observations. 1) The ``support'' posts mostly play an important role in the propagation graph with True Rumor as the target class. 2) The ``deny'' posts contribute more to the propagation graph with False Rumor as the target class. 3) The model with true rumor as the target can capture ``$support \to support$'' propagation pattern ending with $t_6$ as shown in Figure~\ref{fig:CaseStudyTrue}. 4) The model with False rumor as target tends to capture the ``$deny \to support$'' propagation pattern, which ends with $t_4$ as shown in Figure~\ref{fig:CaseStudyFalse}. These observations conform to our assumption and intuition that the propagation structure is vital in the joint task of rumor and stance detection.

\subsubsection{Case Study for Aggregation Model}

\begin{figure}[t!]
\centering
\vspace{-0.1cm}
\setlength{\abovecaptionskip}{3pt} 
\setlength{\belowcaptionskip}{0pt}
\subfigure[RumorEval2019-S (16 MIL classifiers)]{
\includegraphics[scale=0.1]{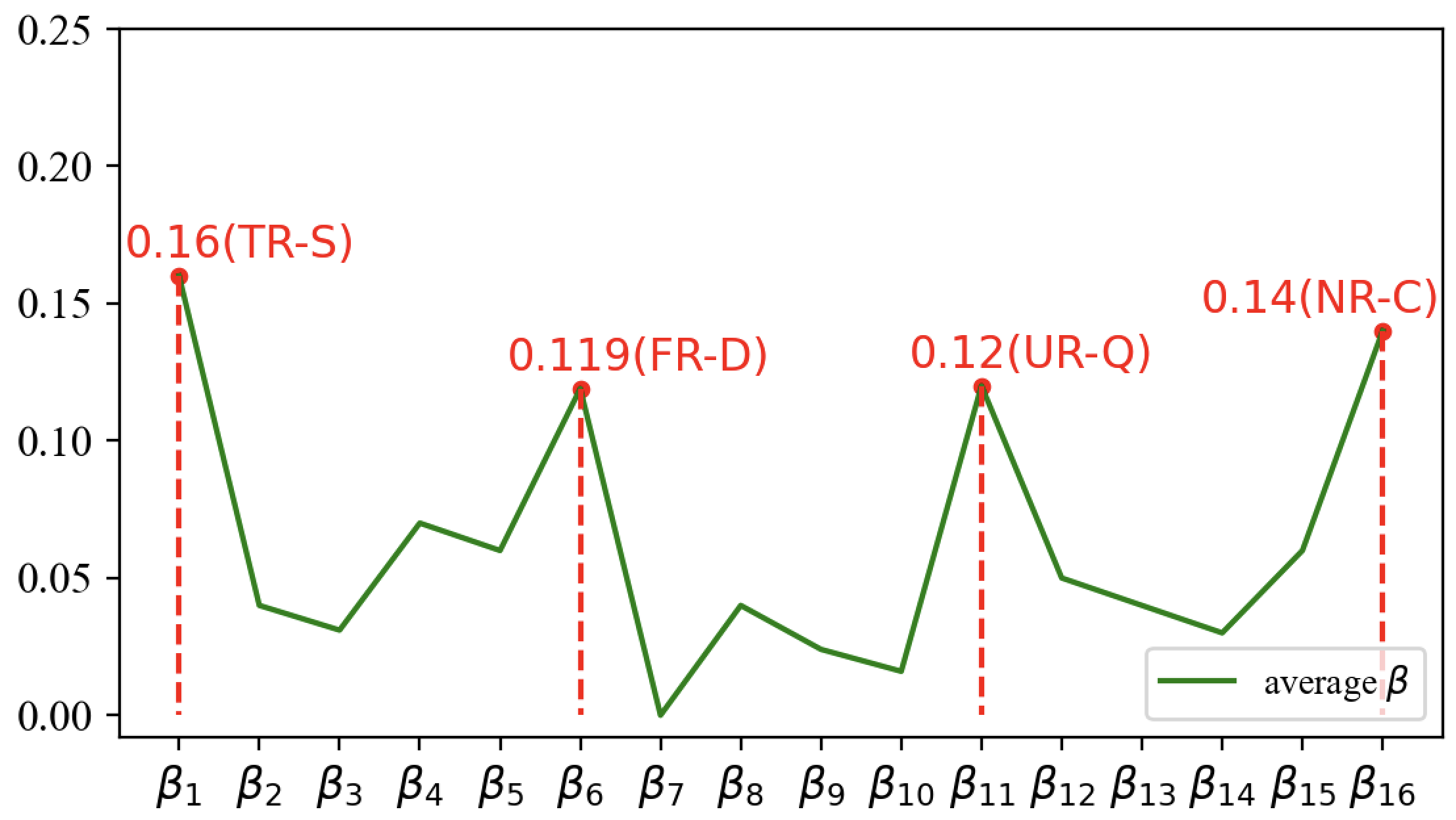}
\label{fig:RumorEval2019-SAttention}}
\subfigure[SemEval8 (12 MIL classifiers)]{
\includegraphics[scale=0.1]{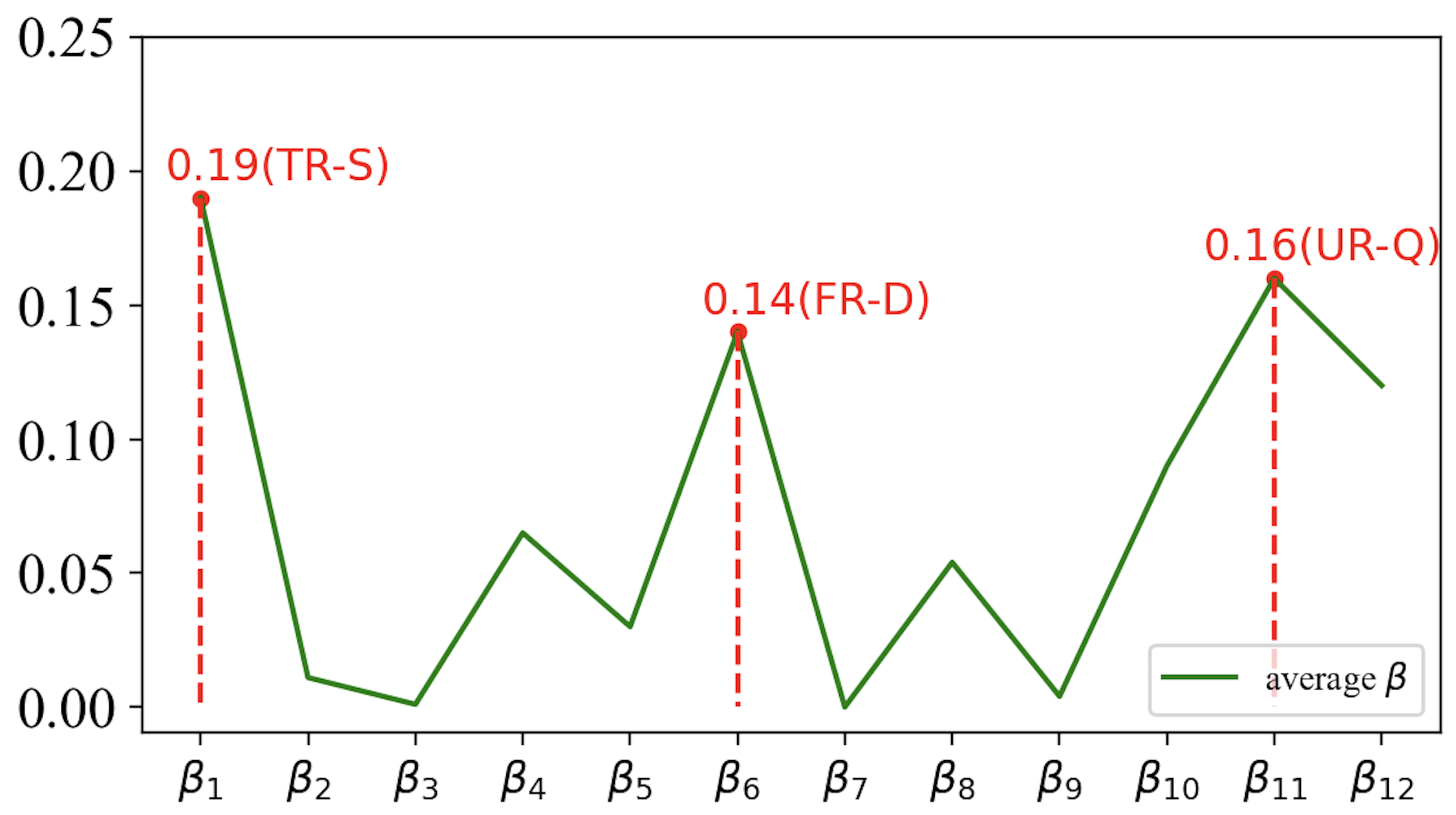}
\label{fig:SemEval8Attention}}
\caption{Average attention scores for binary classifiers obtained from Equation~\ref{equ:betaattention}.}\label{fig:CaseStudy_AggregationAtt}
\vspace{-0.3cm}
\end{figure}

We also conduct experiments to show why the aggregation model can simultaneously boost rumor and stance detection tasks. We randomly sample 100 claims from the PHEME dataset and then disclose the attention scores of all the binary classifiers obtained during the evaluations on the RumorEval2019-S and SemEval8 datasets. The average attention scores over all the claims are shown in Figure~\ref{fig:CaseStudy_AggregationAtt}. 

We have the following observations. 1) The top attention scores indicate a close correlation between the specific rumor veracity and stance category, which is compatible with previous findings~\cite{mendoza2010twitter}. For example, the high values of $\beta_1$ and $\beta_6$ in Figure~\ref{fig:RumorEval2019-SAttention} suggest that True Rumor and Supportive stance in the posts are more closely correlated, and likewise the False Rumor and Denial stance. 
2) The classifiers with lower attention suggest that the final prediction is less affected by the corresponding veracity-stance relationship, indicating that there is a weak correlation between rumor and stance for the current target class. For example, $\beta_2$ and $\beta_3$ in Figure~\ref{fig:SemEval8Attention} demonstrates that for the T-D and T-Q target pairs, the veracity-stance correlation has a weak influence on the prediction of the corresponding binary models. This seems to be consistent with the lower proportion of Denying and Questioning posts as shown in Table~\ref{tab:TwitterData}. 3) The rumor veracity can be generally better determined based on a combination of various stances instead of one-sided stances. For instance, Support and Comment stances combined can contribute more than others when True Rumor is the target class, which can be reflected by the relatively high value when combining T-S ($\beta_1$) and T-C ($\beta_4$) model predictions as shown in Figure~\ref{fig:RumorEval2019-SAttention}. 4) Similarly, among the classifiers with Question stance as the target class, the T-Q  model is generally less important than the other three (i.e., N-Q, F-Q, and U-Q), which indicates the lower proportion of Questioning posts in the True Rumor.\looseness=-1

\vspace{-0.1mm}

\subsection{\ruichao{Error Analysis}}
\begin{figure*}[t!]
\centering
\setlength{\abovecaptionskip}{3pt} 
\setlength{\belowcaptionskip}{0pt}
\includegraphics[width=4in]{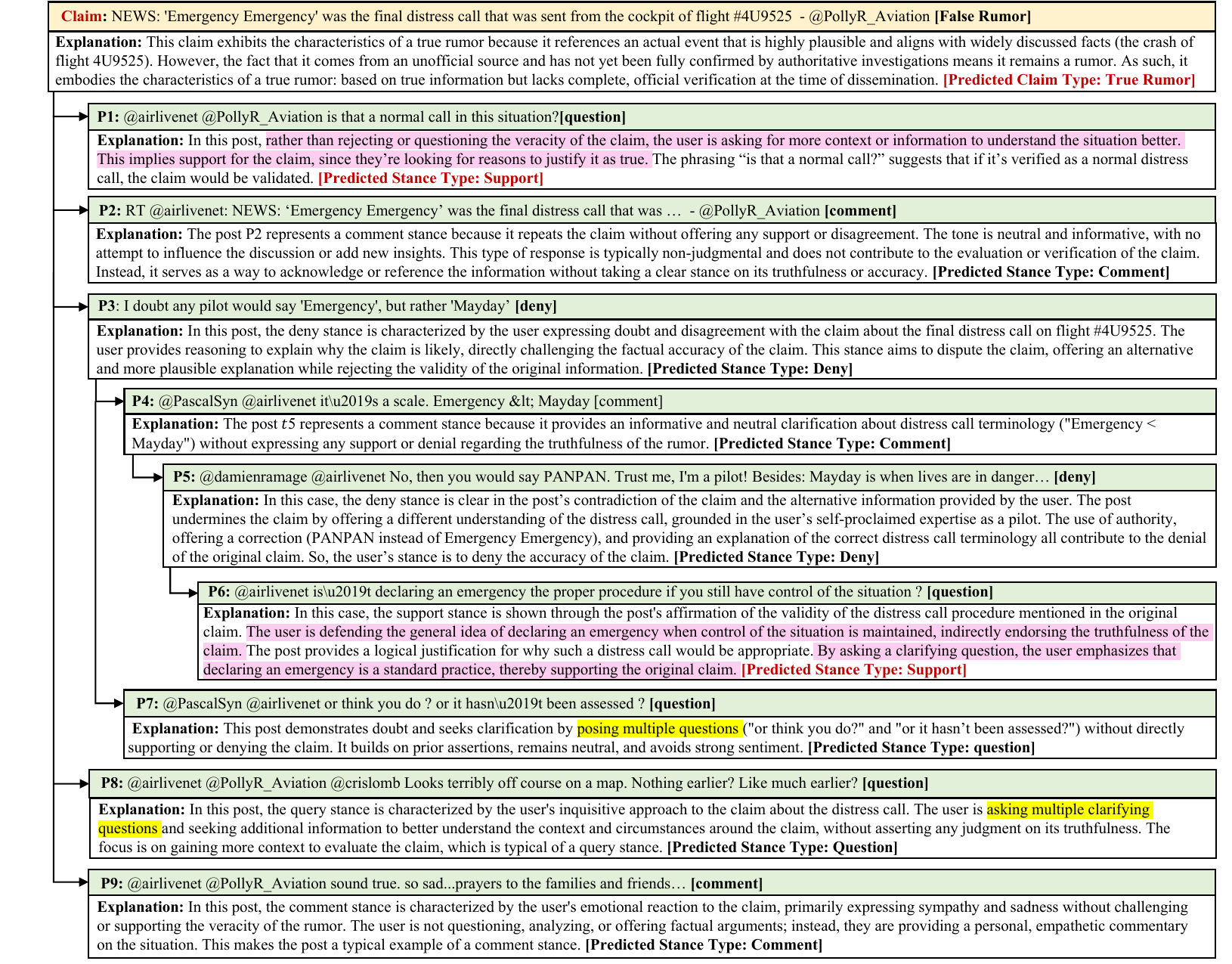}
\caption{\ruichao{Error analysis of LLM-MIL. The yellow box represents the claim, the green box represents the related posts, and the ground truth label of the claim and posts are enclosed in ``[ ]'' at the end of the text. Each claim and post is associated with not only the corresponding explanation generated by LLM but also the claim veracity and stance type predicted by the LLM-MIL.}}
\label{fig:ErrorCase}
\vspace{-0.3cm}
\end{figure*}

\ruichao{To understand the weaknesses of the LLM-MIL and explore the future research direction, we randomly select a misclassified sample from RumorEval2019-S to manually analyze the error the model is prone to make. As Figure~\ref{fig:ErrorCase} shows, a False Rumor claimed that ```Emergency Emergency' was the final distress call that was sent from the cockpit of flight \#4U9525'' has been misclassified as a ``True Rumor''. We analyze this error committed by our model and find that while the explanations generated by LLM for most posts are reliable and most posts can be correctly classified by LLM-MIL, several issues are observed: 1) LLM may hallucinate or fall into extremism when encountering questioning posts. For example, if a user does not deny the source claim but tries to find evidence, it may be considered to support the source claim as highlighted by the pink-color  explanations for P1 and P6. 2) When LLM recognizes multiple questions integrated into one post, it tends to classify such post into questioning stance, as shown by the yellow-color explanations for P7 and P8. Hence, while LLM can generate plausible explanations in many cases, we should pay attention on how it identifies and interprets questioning stance and improve its shortfall on that. }

\subsection{\ruichao{Evaluation of LLM-generated Explanations}}
\ruichao{To evaluate the generated explanations, we randomly sample 10 claims from RumorEval2019-S dataset along with the claim veracity explanations. For brevity and fairness, we randomly select 5 explanation samples for each type of post stances given a claim. Since there is no gold ground-truth for post stance and claim veracity explanation, we provide human-expert explanations for the claims and selected posts. Additionally, we employ the following 5 metrics~\cite{lin2024towards} to evaluate the quality of explanations: 1) Conciseness; 2) Readability; 3) Soundness; 4) Informativeness; 5) Persuasiveness. Although ChatGPT ~\cite{chen2023exploring} can be used for automatic evaluation, our explanations (which are generated by ChatGPT) may be biased towards its preference leading to unfair evaluation results. Hence, we ask 6 annotators to label the explanations and provide their confidence in a 5-point Likert Scale~\cite{joshi2015likert} for each metric.}

\ruichao{Table~\ref{tab:ExplanationEva} exhibits three key observations: 1) Humans can generate concise content with accurate key points, while LLMs often produce overly lengthy explanations. 2) 
The readability and soundness scores for both explanations are comparable, indicating LLMs can generate fluent and logical explanation sentences as human beings for claims and stances. 3) The informativeness and persuasiveness scores for LLM-generated explanations are higher than human beings. This might be because LLMs can capture contextual details precisely beyond human and induce more specific information relevant to the post to make the explanations more persuasive.}

\begin{table}[htbp]
  \centering
  \setlength{\abovecaptionskip}{3pt} 
  \setlength{\belowcaptionskip}{-0.2cm}
  \caption{\ruichao{Evaluation results of the LLM-generated explanations}}
  \resizebox{0.60\textwidth}{!}{
    \begin{tabular}{l|rrrrr}
    \toprule
    \textbf{Explanations} & \multicolumn{1}{l}{\textbf{Conciseness}} & \multicolumn{1}{l}{\textbf{Readability}} & \multicolumn{1}{l}{\textbf{Soundness}} & \multicolumn{1}{l}{\textbf{Informativeness}} & \multicolumn{1}{l}{\textbf{Persuasiveness}} \\
    \midrule
    \textbf{LLM} & 3.76  & 4.35  & 3.67  & 4.13  & 4.12 \\
    \textbf{Human} & 4.16  & 4.36  & 3.63  & 3.81  & 3.67 \\
    \bottomrule
    \end{tabular}%
    }
    \vspace{-0.3cm}
  \label{tab:ExplanationEva}%
\end{table}%

\vspace{-0.2cm}
\section{Discussion}
\ruichao{While we integrate LLM explanations and hierarchical attention mechanisms, our computation overhead does not increase significantly. This is because 1) we only prompt LLM in zero-shot setting to generate explanations for each sample without fine-tuning the model, and 2) the hierarchical stance tree attention mechanism allows for parallel computation across claims. For example, we use a Tesla V100-32GB GPU for training, which just takes 2 hours and 45 minutes for the Twitter15 dataset, and the time elapsed for inference was less than 15 minutes. By leveraging standard optimizations, parallelization, and efficient memory management, our approach can handle the demands of large datasets while providing accurate results. Furthermore, while our model has shown promising results on different datasets, languages, and platforms, there is, to our best knowledge, no generative AI-based rumor dataset in the community, and we plan to collect and analyze such datasets in our future study. }

\section{Conclusion}
Inspired by the multi-instance learning method, we propose a novel weakly supervised framework enhanced by Large Language Model-generated explanations, which is named as LLM-MIL, for detecting rumorous claims and determining the stances of their associated posts simultaneously. Our model is trained solely with bag-level annotations, specifically claim veracity labels. It incorporates a stance aggregation attention mechanism, allowing for the joint inference of both rumor veracity and unseen post-level stance labels. Across both Twitter and Weibo platforms, our approach demonstrates promising results for both rumor and stance detection tasks, outperforming state-of-the-art supervised and unsupervised models. 

\begin{acks}
This work is supported by National Natural Science Foundation of China Young Scientists Fund (No. 62206233) and Hong Kong RGC ECS (No. 22200722).
\end{acks}

\bibliographystyle{ACM-Reference-Format}
\bibliography{sample-base}

\appendix

\end{document}